\documentclass{article} 
\usepackage{iclr2027_conference,times}

\usepackage{amsmath,amsfonts,bm}

\def\eqref#1{equation~\ref{#1}}

\def\1{\bm{1}}

\DeclareMathAlphabet{\mathsfit}{\encodingdefault}{\sfdefault}{m}{sl}
\SetMathAlphabet{\mathsfit}{bold}{\encodingdefault}{\sfdefault}{bx}{n}

\usepackage{hyperref}
\usepackage{url}
\usepackage{booktabs}
\usepackage{array}
\usepackage{enumitem}
\usepackage{amssymb}
\usepackage{graphicx}
\usepackage{capt-of}
\usepackage{multirow}
\usepackage{placeins}
\usepackage[table]{xcolor}
\usepackage{makecell}
\usepackage{colortbl}
\usepackage{wrapfig}
\usepackage{needspace}
\usepackage{pifont}
\usepackage{amsmath}
\usepackage{tabularx}
\usepackage{subfigure}
\usepackage{subcaption}

\newcommand{\ie}{\textit{i.e.}}
\newcommand{\eg}{\textit{e.g.}}

\definecolor{taskwamgainrow}{RGB}{220,235,252}
\definecolor{taskwamgainred}{RGB}{220,0,0}
\newcommand{\taskwamgain}[1]{\textcolor{taskwamgainred}{\textbf{#1}}}

\title{Plausible Future, Wrong Choice: Steering World-Action Models toward Intended Future}

\author{
Zhuohao Li$^{1,2}$ \quad Jiawei He$^3$ \quad Yuxuan Yan$^{4,2}$ \quad Lang Zhou$^{1,2}$ \quad Yinghao Li$^{1,2}$ \\ Yanrui Li$^5$
\quad Tianyu Zhang$^{6,2}$ \quad Yi-Yang Song$^{1,7}$ \quad Xirui-Wu$^8$ \quad Yi-Lin Wei$^{1,9}$\\ Fating Hong$^{10}$ \quad 
Qing Zhang$^1$ \quad Zhilin Zhao$^{1,2}$ \quad Jingdong Wang \quad Wei-Shi Zheng$^{1,2,7,\dagger}$\\[2pt]
\textsuperscript{1}Sun Yat-sen University\\
\textsuperscript{2}Shenzhen Loop Area Institute\\
\textsuperscript{3}The Hong Kong University of Science and Technology (Guangzhou)\\
\textsuperscript{4}Xi'an Jiaotong University\\
\textsuperscript{5}Southern University of Science and Technology\\
\textsuperscript{6}Beijing Institute of Technology\\
\textsuperscript{7}Peng Cheng Laboratory\\
\textsuperscript{8}The Chinese University of Hong Kong Shenzhen
\textsuperscript{9}Ainoro\\
\textsuperscript{10}King Abdullah University of Science and Technology\\
\texttt{lizhh268@mail2.sysu.edu.cn}\\[-1pt]
$^\dagger$ Corresponding authors}

\iclrfinalcopy 
\begin{document}

\maketitle

\begin{figure}[!h]
    \centering
    \includegraphics[width=\linewidth]{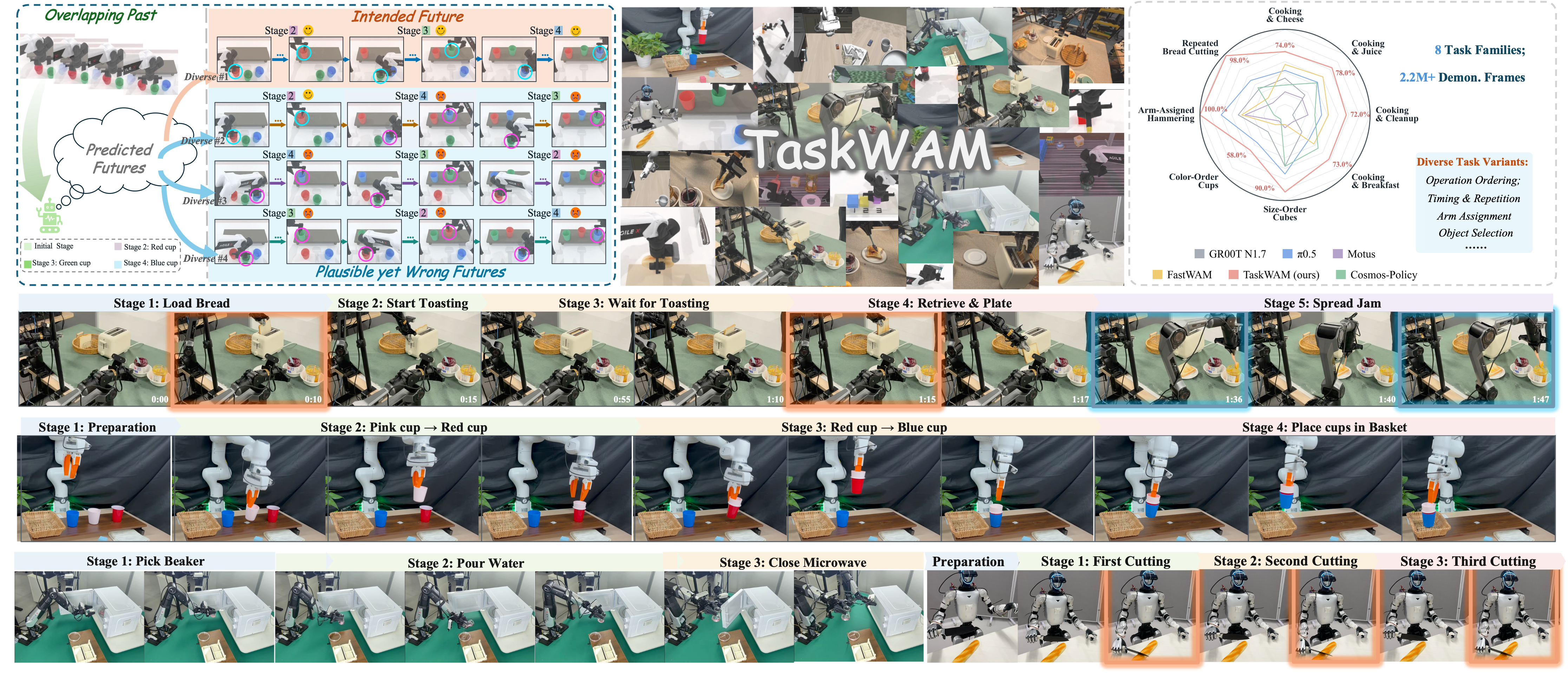}
\caption{
    \textbf{From plausible futures to reliable stage progression.}
    \textbf{Top (left):}
    Overlapping task trajectories can induce Stage Confusion: WAMs may follow plausible yet incorrect execution routes.
    \textbf{Top (right):}
    StageBench systematically evaluates this challenge through controlled overlapping multi-stage trajectories across eight task families and more than 2.2 million demonstration frames.
    \textbf{Bottom:}
    Visualization of TaskWAM performing reliable long-horizon manipulation across diverse robot embodiments and stage progressions.
    Visually similar stages are highlighted with matching color borders to indicate potential stage confusion.
    }
    \label{fig:teaser}
\end{figure}

\begin{abstract}
World-Action Models (WAMs) have emerged as a promising paradigm for embodied generalist policies by jointly modeling future world evolution and robot actions across diverse manipulation tasks.
However, representing diverse futures is only the first step: a capable WAM must also generate futures that match the requirements of the current task and execution progress.
In long-horizon manipulation where similar observation-action segments occur across different tasks and stages,
existing approaches may generate locally plausible yet misdirected rollouts to drift toward wrong task stage, \ie, a phenomenon we term \emph{Stage Confusion}.
To address this problem, we propose \textbf{TaskWAM}, a closed-loop, stage-aware framework that represents fine-grained task requirements with \emph{Mode Tokens} and tracks execution progress with \emph{State Tokens}, forming a structured task state to steer world--action generation toward intended future.
We further introduce \textbf{StageBench}, a controlled multi-stage manipulation benchmark with parameterized stage variations and automated evaluation.
Extensive experiments on both simulation benchmarks and real-world tasks demonstrate that TaskWAM effectively reduces confusion and achieves competitive performance with state-of-the-art methods. 
Moreover, TaskWAM provides strong plug-and-play transfer across heterogeneous WAM architectures and robot embodiments, including single-arm robots, dual-arm grippers, dexterous hands, and humanoid robots.
\end{abstract}   

\begin{center}
Project Page: \url{https://isee-laboratory.github.io/TaskWAM/}
\end{center}

\section{Introduction}
\label{sec:intro}

Recently, World-Action Models (WAMs) \citep{bi2026motus,ye2026dreamzero,li2026lingbotva,yuan2026fastwam,chen2026lawam} have shown promise for embodied generalist policies by jointly modeling future world evolution and robot actions, with the goal of supporting diverse manipulation tasks in everyday environments\citep{zhu2025uwm,li2025uva,li2026lingbotva}.
However, modeling diverse futures is only the first step.
Once models are capable of representing multiple possible futures, the next challenge lies in generating future that is consistent
with the task requirements and execution context.
%
Existing WAMs typically infer stage progression implicitly from task instructions, observations, and execution histories \citep{lin2026jepawamlearningvisionlanguageactionpolicies,zhao2026sgwamselfguidedworldmodeling}.
This becomes particularly challenging in long-horizon household manipulation, where similar observation-action interactions can correspond to different next stages depending on fine-grained task requirements, \eg, ordering, timing, repetition, or embodiment assignment, as illustrated in Figure~\ref{fig:teaser} (\textit{Top left}).

Consequently, WAMs may generate futures that are visually plausible and physically feasible, yet lead execution to the wrong stage.
We refer to this failure as \emph{stage confusion in world-action modeling} (hereafter, \emph{stage confusion}).
A few recent WAMs \cite{yang2026memorywam,cai2026ahawam} have explored longer-term history modeling, but stage confusion is not merely a memory-length problem, leaving two key challenges:
(1) similar interaction histories may still require different next stages, making instruction-specific guidance essential for resolving stage ambiguity; and
(2) simply extending history introduces substantial redundant context, which can obscure critical execution cues for stage tracking while increasing the architectural burden of WAMs.

To address these challenges, we propose \textbf{TaskWAM}, a closed-loop, stage-aware framework that explicitly grounds world-action generation in both task requirements and execution progress for long-horizon manipulation.
Specifically, we introduce \emph{Structured Task-State Grounding} with two complementary representations: 
\emph{Mode Tokens} retrieve mode representations from a learnable mode codebook and ground them with text context to encode fine-grained execution requirements, while \emph{State Tokens} track execution progress through causal temporal modeling of  closed-loop  action-observation transitions.
We further propose \emph{Task-State Conditioned Future Steering}, which steers future prediction by inject the structured task state into world and action generation through attention conditioning and residual flow correction.
TaskWAM thus bridges diverse future modeling and intended future generation, offering two main advantages: 
(i) future generation grounding in closed-loop robot-environment interaction; 
(ii) plug-and-play steering mechanism applicable across heterogeneous WAM backbones.

Additionally, existing benchmarks \citep{mees2022calvin,NEURIPS2023_8c3c6668,chen2025robotwin2,ICLR2026_a05003fd} has mainly evaluated tasks in relatively independent settings, lacking sufficient data and evaluation protocols to assess this capability.
To support systematic evaluation, we further introduce \textbf{StageBench}, a controlled multi-stage manipulation benchmark built on LIBERO \cite{NEURIPS2023_8c3c6668} and RoboTwin \cite{chen2025robotwin2}.
StageBench constructs overlapping observation-action trajectories across tasks and execution stages by varying stage compositions and orders, covering visual grounding, sequential reasoning, embodiment coordination, \textit{etc.}
Extensive experiments in simulation and diverse real-world robotic tasks demonstrate that TaskWAM significantly reduces stage confusion and improves multi-stage manipulation over state-of-the-art methods.
We further validate its generalization on RoboTwin 2.0 benchmark \cite{chen2025robotwin2} and its plug-and-play applicability across heterogeneous WAM architectures.
Moreover, our method can be applied across diverse robot embodiments, including single-arm robots, dual-arm grippers, dexterous hands, and humanoid platforms.
Our contributions are summarized as follows:
\begin{itemize} 
\item 
We identify \emph{stage confusion in world-action modeling}, where WAMs generate locally plausible yet stage-inconsistent futures under overlapping observation-action trajectories.
\item  
We propose TaskWAM, which explicitly grounds world-action generation in task requirements and execution progress through Structured Task-State Grounding and task-state-conditioned future steering for stage-consistent future generation.
\item  
We introduce StageBench, a controlled multi-stage manipulation benchmark that systematically constructs overlapping trajectories through parameterized task-stage variations and provides automated stage-aware evaluation.
\item Extensive experiments across simulation and real-world tasks demonstrate that TaskWAM significantly reduces stage confusion and improves multi-stage manipulation across heterogeneous WAM backbones and diverse robot embodiments.
\end{itemize}

\section{Related Work}
\label{sec:related_work}


\noindent\textbf{World-Action Models.}
Generalist robot learning aims to develop unified policies that can handle diverse manipulation tasks within a single model.
Vision-Language-Action models \citep{physicalintelligence2025pi05,nvidia2026gr00tn17} have advanced this direction by scaling language-conditioned robot policies.
World-Action Models (WAMs) \citep{bi2026motus,zhu2025uwm,ye2026dreamzero,li2026lingbotva,yuan2026fastwam, cai2026ahawam,chen2026lawam} further extend this paradigm by jointly modeling future observations and robot actions. 
Recent studies explores inference-time imagination, world-action information exchange, and more efficient generation schedules \citep{ye2026dreamzero,li2026lingbotva,yuan2026fastwam}, such as DreamZero \citep{ye2026dreamzero} and LingBot-VA \citep{li2026lingbotva}.
However, existing WAMs still typically learn world-action futures from unified conditioning contexts \citep{lin2026jepawamlearningvisionlanguageactionpolicies,zhao2026sgwamselfguidedworldmodeling}, leaving stage progression to be inferred implicitly from instructions, observations, and histories.
Consequently, a predicted future can remain locally plausible while leading the robot to the wrong stage, \ie, \emph{plausible future, yet wrong choice}.
In contrast, TaskWAM explicitly grounds future generation in task requirements
and execution progress, targeting stage confusion in long-horizon manipulation.

\noindent\textbf{Historical Context Modeling.}
Temporal context has been widely used to support long-horizon decision making
in language modeling \citep{dai-etal-2019-transformer},
navigation \citep{Fang_2019_CVPR},
video generation \citep{Henschel_2025_CVPR},
and robot manipulation \citep{pmlr-v205-guhur23a}.
Existing robot policies incorporate history through
temporal abstraction \citep{Belkhale-RSS-24},
visual memory \citep{pmlr-v267-fang25c},
retrieval \citep{ShahR-RSS-26},
or explicit state representations \citep{chung2026memorystate}.
However, task instructions and longer histories is insufficient to guarantee that
task requirements remain effective throughout execution; policies may still
rely on easier visual-action correlations \citep{lin2026jepawamlearningvisionlanguageactionpolicies,zhao2026sgwamselfguidedworldmodeling}.
Moreover, longer execution contexts often contain redundant information,
which can dilute critical execution cues while increasing design complexity and
computational cost \citep{ICLR2026_9da515b1}.
Rather than extending raw history, TaskWAM extracts progress-relevant evidence
from causal action-observation transitions for explicit execution tracking.

\noindent\textbf{Long-Horizon Multi-Stage Manipulation.}
Long-horizon household tasks are common in daily life and remain a key challenge for reliable robot deployment \citep{pmlr-v205-li23a}.
A common characteristic of these tasks is that similar observation-action interactions often recur within the same scene across different tasks and execution stages \citep{ICLR2026_a05003fd}.
Reliable execution therefore requires robots to distinguish such overlapping trajectories and complete the required stages in demanding order~\citep{mees2022calvin,pmlr-v205-li23a}.
However, prior work has largely learned and evaluated tasks in relatively isolated settings \citep{OpenX24}, leaving this capability underexplored.
In this work, we target this setting with a stage-aware approach and introduce a controlled multi-stage benchmark for systematic evaluation.
\section{TaskWAM Framework}
\label{sec:method}

\subsection{Problem Formulation}
\label{sec:problem_setup}

\noindent\textbf{Stage Confusion.}
For task $i$, we denote a rollout by $\mathcal{T}_i=\{\ell_i,\xi_i\}$, where $\xi_i=\{(o_t^i,a_t^i)\}_{t=1}^{T_i}$ is segmented into a stage sequence $\{s_i^1,\ldots,s_i^{n_i}\}$.
At the transition from stage $k$, let $\mathcal{H}_i^k$ denote the observed execution history and $s_i^{k+1}$ the subsequent stage.
Two stage-transition points $(i,k)$ and $(j,k')$ may have similar local histories, $\mathcal{H}_i^k\approx\mathcal{H}_j^{k'}$, but require different subsequent stages, $s_i^{k+1}\neq s_j^{k'+1}$.
WAMs may generate a locally plausible future $\hat{s}_i^{k+1}$ yet advances execution to the wrong stage, which we refer to as \emph{Stage Confusion}. 
It typically occurs in two cases:
\ding{182} \emph{Cross-task confusion}, where the model continued toward a wrong stage from another task, \ie, $\hat{s}_i^{k+1}=s_j^{k'}$, $j\neq i$;
\ding{183} \emph{Within-task confusion}, where the model mistakes similar stages within the same task for one another, leading to stage repetition or skipping, \ie, $\hat{s}_i^{k+1}=s_i^{k'}$, $k'\neq k+1$.

\noindent\textbf{Task-State Grounding for Future Generation.}
Stage progression depends on two complementary task factors: 
task requirements specify  how execution should proceed, while closed-loop interaction history provides evidence of the progress achieved so far.
We represent them as an explicit task state $\mathbf{T}_t=[\mathbf{M};\mathbf{S}_t]$,
where $\mathbf{M}$ encodes fine-grained task requirements grounded in instruction $\ell$, and $\mathbf{S}_t$ tracks execution progress from history $\mathcal{H}_t$.
We condition future world evolution $\mathbf{W}_t^{+}$ and
robot actions $\mathbf{A}_t^{+}$ on this structured state:
\begin{equation}
    (\widehat{\mathbf{W}}_t^{+},\widehat{\mathbf{A}}_t^{+})
    \sim
    p_{\theta}
    \bigl(
        \mathbf{W}_t^{+},\mathbf{A}_t^{+}
        \mid \ell,o_t,\mathcal{H}_t,\mathbf{T}_t
    \bigr).
    \label{eq:task_state_generation}
\end{equation}
Our goal is to explicitly model stage progression in world-action generation rather than leaving it to be implicitly inferred from an unstructured conditioning context.

\begin{figure}[t]
    \centering
    \includegraphics[width=\linewidth]{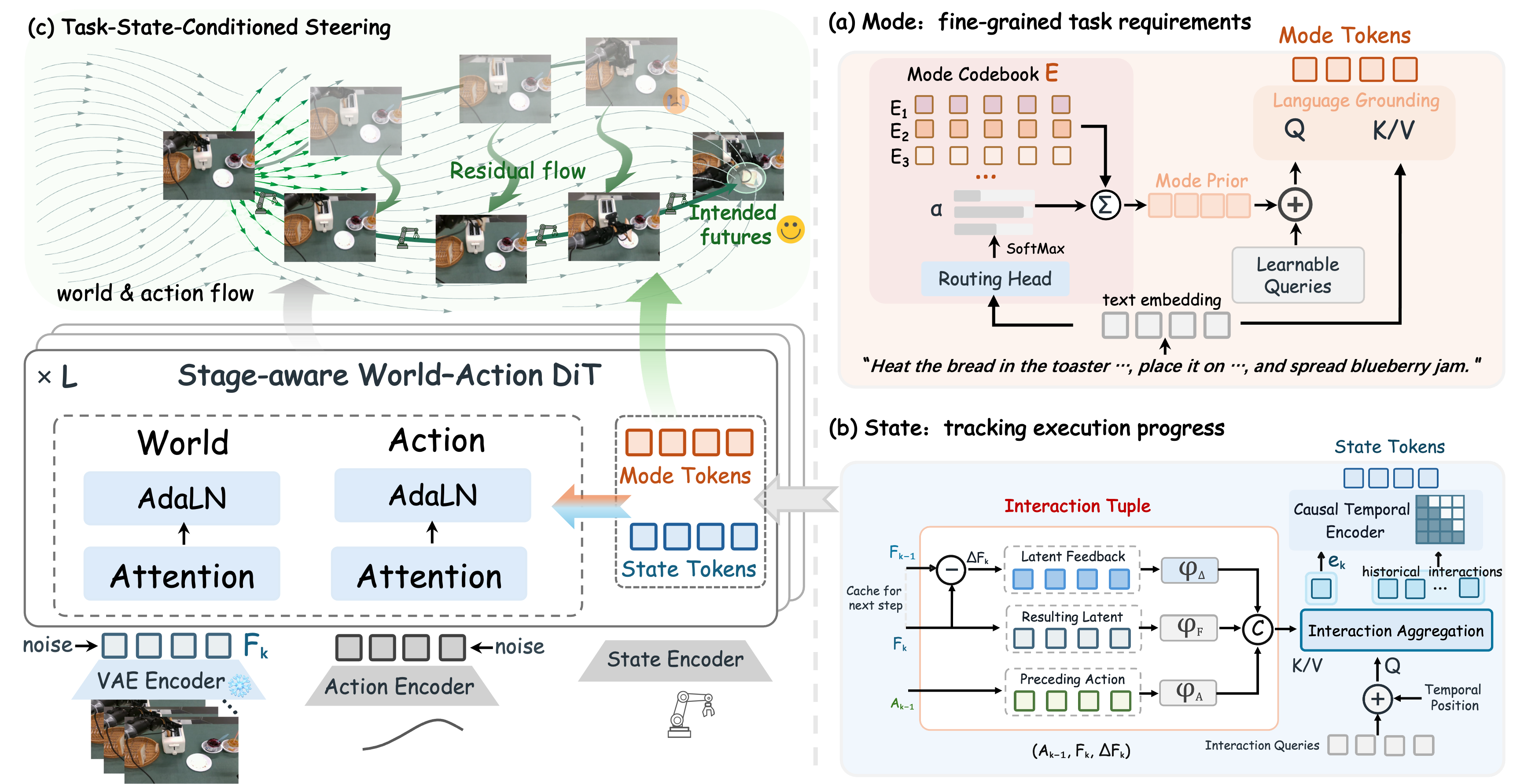}
    \caption{
    \textbf{Overall framework.} TaskWAM augments a World-Action DiT through explicit task-state grounding and conditioning.
    \textbf{(a) Mode Tokens} retrieve mode representations from a learnable codebook and ground them with text context to encode fine-grained task requirements.
    \textbf{(b) State Tokens} aggregate action-observation interactions into a causal interaction history that tracks execution progress.
    \textbf{(c) Task-state-conditioned steering} conditions the world and action streams through attention and modulation, while residual flow guidance steers generation toward  intended futures.}
    \vspace{-4mm}
    \label{fig:taskwam_overview}
\end{figure}

\subsection{Overview}
TaskWAM grounds world--action generation in task requirements and
execution progress to reduce stage confusion.
As illustrated in Figure~\ref{fig:taskwam_overview}, it consists of two core components: 
(1) \emph{Structured Task-State Grounding} constructs Mode Tokens through codebook retrieval and text grounding, and State Tokens through interaction aggregation and causal temporal encoding; and
(2) \emph{Task-State Conditioned Future Steering} incorporates the resulting task state into backbone conditioning and velocity predictions, steering generation toward the intended future.

\subsection{Structured Task-State Grounding}
\label{sec:task_state}
%
Following Sec.~\ref{sec:problem_setup}, we construct Mode Tokens $\mathbf{M}\in\mathbb{R}^{N_m\times d}$ and State Tokens $\mathbf{S}_t\in\mathbb{R}^{N_s\times d}$ to form the structured task state
$\mathcal{T}_t=[\mathbf{M};\mathbf{S}_t]$.

\noindent\textbf{Mode Tokens: instruction-grounded requirements.}
Tasks can share the same manipulation steps while differing in fine-grained requirements, \eg, ordering, timing, repetition, and embodiment assignment~\citep{ICLR2025_49f80e4d}.
To keep these distinctions available throughout execution, we retrieve mode representations from a learnable codebook and ground them in the full instruction context.

Given instruction $\ell$, the pretrained T5 encoder~\citep{JMLR:v21:20-074} produces text embedding $\mathbf{C}=\mathcal{E}_{\mathrm{text}}(\ell) \in\mathbb{R}^{L\times D}$.
A routing head then predicts a soft assignment over $K_{\mathrm{prior}}$ learnable modes:
\begin{equation}
    \mathbf{c}_{\mathrm{pool}}
    =
    \frac{
        \sum_{i=1}^{L} \mathrm{mask}_{i}\mathbf{c}_{i}
    }{
        \sum_{i=1}^{L} \mathrm{mask}_{i}
    },
    \qquad
    \boldsymbol{\alpha}
    =
    \mathrm{softmax}
    (
        W_2\,
        \mathrm{SiLU}
        (W_1\mathbf{c}_{\mathrm{pool}}+b_1)
        +b_2
    ),
    \label{eq:mode_routing}
\end{equation}
where $\mathrm{mask}_i$ identifies valid tokens and $\boldsymbol{\alpha}\in\mathbb{R}^{K_{\mathrm{prior}}}$.

The mode codebook $\mathbf{E}\in\mathbb{R}^{K_{\mathrm{prior}}\times N_m\times D}$ stores $N_m$ learnable slots per mode. 
We retrieve the corresponding mode slots through soft routing $\mathbf{m}_{l}^{\mathrm{raw}}
=\sum_{k=1}^{K_{\mathrm{prior}}}\alpha_k\mathbf{E}_{k,l}$ for $l=1,\ldots,N_m$.
The retrieved slots provide a coarse execution-requirement prior, serving as a prior for constructing Mode Tokens.
We then use these slots as queries over the original language-token sequence to incorporate instruction-specific requirements, such as the precise ordering and timing:
\begin{equation}
    \mathbf{m}_{l}
    =
    \mathcal{Q}_{\mathrm{ground}}
    \left(
        \mathbf{m}_{l}^{\mathrm{raw}},
        \mathbf{C}
    \right),
    \qquad
    \mathbf{M}
    =
    [\mathbf{m}_{1},\ldots,\mathbf{m}_{N_m}]
    \in
    \mathbb{R}^{N_m\times d}.
    \label{eq:mode_grounding}
\end{equation}
where $\mathcal{Q}_{\mathrm{ground}}$ performs language cross-attention and projects its output to dimension $d$.
The resulting Mode Tokens combine the retrieved prior with detailed task requirements and are retained across stages, providing persistent task guidance throughout world-action generation.

\noindent\textbf{State Tokens: causal temporal execution tracking.}
Mode Tokens specify the task requirements but do not indicate which stages have been completed.
We therefore construct State Tokens from actions and their observed outcomes in closed-loop robot-environment interaction, rather than encoding raw visual history alone.
A causal execution-state encoder first aggregates each interaction into a compact token and then models the sequence over time.

Specifically, for each history slot $k$, we pair the preceding action chunk $\mathbf{A}_{k-1}$ with the visual latent $\mathbf{F}_k$ of the resulting observation, encoded by the Wan Video VAE~\citep{wan2025}, and its change
$\Delta\mathbf{F}_k=\mathbf{F}_k-\mathbf{F}_{k-1}$.
After modality-specific projection and positional encoding, learnable queries $\mathbf{q}_k$ aggregate these signals through a Q-Former~\citep{pmlr-v202-li23q}:
\begin{equation}
    \mathbf{e}_k=
    \mathcal{Q}_{\mathrm{s}}
    \left(
        \mathbf{q}_k,\,
        [
        \phi_v(\mathbf{F}_k);
        \phi_\Delta(\Delta\mathbf{F}_k);
        \phi_a(\mathbf{A}_{k-1})
        ]
    \right)
    \in\mathbb{R}^{d}.
    \label{eq:placekeeping_transition}
\end{equation}
where $\phi_v$, $\phi_\Delta$, and $\phi_a$ are modality-specific projections.
A causal temporal encoder then processes the most recent $K$
interaction tokens in execution order:
\begin{equation}
    \mathbf{S}_t
    =
    \mathcal{E}_{\mathrm{causal}}
    \left(
        \mathbf{e}_{t-K+1:t}
    \right)
    \in\mathbb{R}^{K\times d}.
    \label{eq:state_tokens}
\end{equation}
with $N_s=K$.
The positional encodings identify interaction order, while the causal attention mask restricts each slot to current and preceding interactions, allowing State Tokens to summarize how interactions unfolded over time.

During deployment at step $t$, the robot applies the predicted actions $\mathbf{A}_t$ and encodes the next observation $\mathbf{F}_{t+1}$, the new transition $(\mathbf{F}_{t+1},\Delta\mathbf{F}_{t+1},\mathbf{A}_t)$ is then appended, and $\mathbf{S}_{t+1}$ is updated.
This supplies subsequent world--action generation with progress information from robot--environment feedback.
Detailed settings of both token constructions are provided in
Appendix~\ref{app:state_encoder}.

\subsection{Task-State Conditioned Future Steering}
\label{sec:tffs}
Given the structured task state $\mathbf{T}_t$, TaskWAM steers world--action generation toward the intended future through backbone conditioning and residual velocity correction.

\noindent\textbf{Task-State Conditioning.}
We append $\mathbf{T}_t$ to the key-value context of both streams.
World and action tokens can therefore attend to the same task requirements and execution progress throughout generation.
To keep the active execution mode effective across backbone layers, we further pool the Mode Tokens into $\bar{\mathbf{m}}\in\mathbb{R}^{d}$ and apply
layer-wise AdaLN modulation~\citep{Peebles_2023_ICCV}:
\begin{equation}
    [\boldsymbol{\gamma}^{m}_{l},
     \boldsymbol{\beta}^{m}_{l}]
    =
    g_l(\bar{\mathbf{m}}),
    \qquad
    \widetilde{\mathbf{H}}_{l}
    =
    (1+\boldsymbol{\gamma}^{m}_{l})
    \odot
    \mathrm{LN}(\mathbf{H}_{l})
    +
    \boldsymbol{\beta}^{m}_{l},
    \label{eq:mode_adaln}
\end{equation} 
where $g_l$ denotes a layer-specific modulation network. 
This carries instruction-specific requirements across generation layers while attention incorporates the updated execution state.
The design preserves the backbone's native world-action attention relations, supporting integration across heterogeneous WAM architectures (Sec.~\ref{sec:main_results}).

\noindent\textbf{World-Action Flow Steering.}
Beyond backbone conditioning, we apply task-state-conditioned residual corrections~\citep{NEURIPS2025_300ccb21} to steer world and action generation toward the intended future.
For flow-based backbones, let $\mathbf{H}_{t,\tau}^{b}$ and $\mathbf{v}_{t,\tau}^{b}$ denote the hidden states and velocity predictions at rollout step $t$ and flow time $\tau$, where $b\in\{w,a\}$ denotes the world or action stream.
Each lightweight cross-attention head $\mathcal{R}_b$ uses the
corresponding hidden states to query $\mathbf{T}_t$ and predict
a velocity residual:
\begin{equation}
\begin{aligned}
    \Delta\mathbf{v}_{t,\tau}^{b}
    &=
    \mathcal{R}_{b}
    (\mathbf{H}_{t,\tau}^{b};\mathbf{T}_t),
    \qquad 
    \widehat{\mathbf{v}}_{t,\tau}^{b} =
    \mathbf{v}_{t,\tau}^{b}
    +
    \gamma_b\Delta\mathbf{v}_{t,\tau}^{b},
\end{aligned}
\label{eq:tffs_flow}
\end{equation}
where $\gamma_b$ is a learnable bounded scale.
Both flows are corrected with the same task state, steering the intended stage progression rather than merely producing
mutually consistent world and action predictions.
For Cosmos Policy, the residual instead corrects denoised latent predictions under its native EDM formulation.

\noindent\textbf{Training Objective.}
We train TaskWAM end-to-end using each backbone's vanilla generative objective, \ie,
flow matching for flow-based WAMs and EDM denoising for Cosmos Policy.
The overall training objective is
$    \mathcal{L}
    =
    \mathcal{L}_{\mathrm{world}}
    +
    \mathcal{L}_{\mathrm{action}}$.
More implementation details are provided in Appendix~\ref{app:training_inference}.
\section{Stage-Confusion Benchmark}
\label{sec:stage_bench}
To support the systematic evaluation of stage confusion, we construct \emph{StageBench} on the single-arm LIBERO and bimanual RoboTwin 2.0 platforms.
StageBench includes diverse kitchen and household scenarios with meaningful multi-stage manipulation routines, comprising 10 task families with visual grounding, sequential reasoning, temporal reasoning, and embodiment coordination, \textit{etc.}, as illustrated in Fig.~\ref{fig:stage_bench}. 
Additional details are provided in Appendix~\ref{app:stage_bench}.

\begin{figure*}[t]
    \centering
    \includegraphics[width=0.98\textwidth]{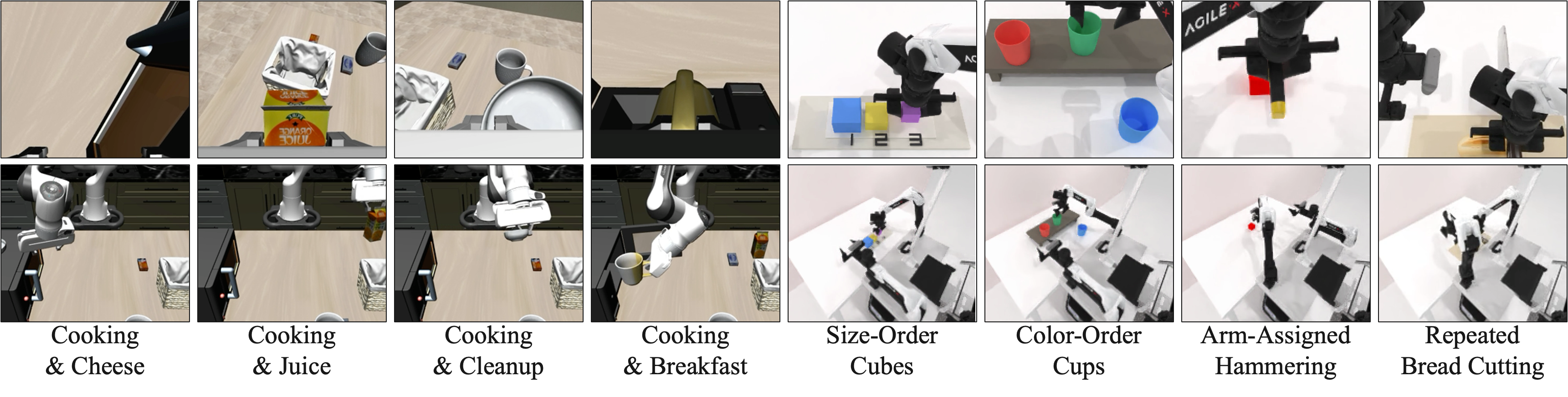}
    \vspace{-2mm}
    \caption{The visualization of task families in StageBench (see Appendix ~\ref{app:stage_bench} for more details).}
    \vspace{-2mm}
    \label{fig:stage_bench}
\end{figure*}

\noindent\textbf{Task design and data collection.}
Each task family is instantiated into multiple \emph{task variants} built from the same scene, objects, and manipulation stages, with matched stage counts and comparable difficulty.
Across variants, we construct different stage sequences by changing their order, timing, arm assignment, or repetition to target \emph{cross-task confusion}.
Within each variant, repeated or similar interaction stages appear at different execution positions to target \emph{within-task confusion}.
We collect demonstration trajectories through 3D mouse teleoperation and automated data generation, resulting in over 2.20 million frames in total, with approximately 12 hours of manual teleoperation.

\noindent\textbf{Evaluation.}
We build an automatic evaluator that reconstructs stage-wise execution from simulator states.
During each rollout, it records task-relevant object states, contacts, arm actions, and their timestamps, and matches these events against the stage definitions of the corresponding task variant.
This allows us to summarize which stages are entered and completed throughout execution, from which we derive task completion, execution route, and stage-level progress.
For each task, we manually inspect 100 evaluation results to confirm the reliability of the automatic evaluator.
\section{experiment}
\label{sec:experiment}
\subsection{Experimental Setup}
\noindent\textbf{Simulation Benchmarks.}
We evaluate our method in two robotic simulation benchmarks:
(1) In \textbf{StageBench}, a controlled multi-stage benchmark with 8 task families spanning single-arm and bimanual manipulation, as detailed in Sec.~\ref{sec:stage_bench}.
(2) In \textbf{RoboTwin 2.0}~\cite{chen2025robotwin2}, we follow the same experimental settings of \citep{bi2026motus,yuan2026fastwam,li2026lingbotva}, training all models on 2,500 demonstrations collected in clean scenes (50 per task) and 25,000 demonstrations collected in randomized scenes (500 per task).

\noindent\textbf{Baselines.}
We compare our approach against the state-of-the-art (SOTA) baselines,
including WAMs:
Motus~\citep{bi2026motus},
Cosmos Policy~\citep{kim2026cosmospolicy},
DreamZero~\citep{ye2026dreamzero},
Fast-WAM~\citep{yuan2026fastwam},
LingBot-VA~\citep{li2026lingbotva},
LaWAM~\citep{chen2026lawam},
and AHA-WAM~\citep{cai2026ahawam}.
We further include classical VLA baselines, including
$\pi_{0.5}$~\citep{physicalintelligence2025pi05}
and GR00T N1.7~\citep{nvidia2026gr00tn17}.
We evaluate both final task success and stage-aware progression,
including stage-wise completion and Stage Progress Rate (\textit{SPR}), to
measure reliable multi-stage execution beyond overall completion.

\noindent\textbf{Implementation Details.}
We build our framework upon Fast-WAM~\citep{yuan2026fastwam} and Cosmos Policy~\citep{kim2026cosmospolicy}, preserving their native inference structures while injecting task-state conditioning. 
For StageBench, all methods are trained and evaluated under the same data and evaluation settings.
For RoboTwin 2.0, we follow its standard training and evaluation protocols. 
Experiments are trained on $16\times$ NVIDIA H100 80GB GPUs, and all simulation results are evaluated over 100 trials.
More details are provided in the supplementary material.

\begin{table*}[t]
\caption{\textbf{Performance comparison on diverse long-horizon manipulation tasks of StageBench.}
Our method outperforms all baselines in Success Rate (\textit{Suc.}, \%$\uparrow$).
Best results are in \textbf{bold}.}
\label{tab:stagebench_main_results}
\centering
\small
\resizebox{0.95\textwidth}{!}{%
\begin{tabular}{@{}lccccccccc@{}}
\toprule
\multirow{3}{*}{\textbf{Method}}
& \multicolumn{4}{c}{\textbf{Kitchen Manipulation}}
& \multicolumn{4}{c}{\textbf{Bimanual Manipulation}}
& \multirow{3}{*}{\textbf{\shortstack{Avg.\\ \textit{Suc.}}}} \\
\cmidrule(lr){2-5}
\cmidrule(lr){6-9}

& \shortstack{Cooking\\\& Cheese}
& \shortstack{Cooking\\\& Juice}
& \shortstack{Cooking\\\& Cleanup}
& \shortstack{Cooking\\\& Breakfast}
& \shortstack{Size-Order\\Cubes}
& \shortstack{Color-Order\\Cups}
& \shortstack{Arm-Assigned\\Hammering}
& \shortstack{Repeated\\Bread Cutting}
& \\
\midrule

GR00T N1.7
& 43 & 33 & 9 & 28 & 60 & 28 & 55 & 43
& 37.4 \\

$\pi_{0.5}$
& 52 & 54 & 43 & 40 & \textbf{69} & 46 & 75 & \textbf{57}
& 54.5 \\

Motus
& 27 & 28 & 27 & 12 & 15 & 10 & 47 & 20
& 23.3 \\

\midrule
Cosmos Policy
& 35 & 53 & 35 & 55 & 18 & 7 & 45 & 13
& 32.6 \\

\textbf{w/ TaskWAM}
& 57 & 61 & 45 & 63 & 29 & 22 & 56 & 31
& 45.5 \\

\rowcolor{taskwamgainrow}
$\Delta$
& \taskwamgain{+22} & \taskwamgain{+8} & \taskwamgain{+10} & \taskwamgain{+8} & \taskwamgain{+11} & \taskwamgain{+15} & \taskwamgain{+11} & \taskwamgain{+18}
& \taskwamgain{+12.9} \\

\midrule
FastWAM
& 59 & 63 & 51 & 48 & 11 & 16 & 51 & 28
& 40.9 \\

\textbf{w/ TaskWAM}
& \textbf{74} & \textbf{78}
& \textbf{72} & \textbf{73}
& 65 & \textbf{54}
& \textbf{83} & 56
& \textbf{69.4} \\

\rowcolor{taskwamgainrow}
$\Delta$
& \taskwamgain{+15}
& \taskwamgain{+15}
& \taskwamgain{+21}
& \taskwamgain{+25}
& \taskwamgain{+54}
& \taskwamgain{+38}
& \taskwamgain{+32}
& \taskwamgain{+28}
& \taskwamgain{+28.5} \\

\bottomrule
\end{tabular}%
}
\vspace{-2mm}
\end{table*}
\begin{table*}[t]
\caption{\textbf{Performance comparison on RoboTwin 2.0 Benchmark (Clean vs. Randomized, 50 tasks).} We show the representative tasks and full comparison results is in Appendix~\ref{app:robotwin50}.}
\label{tab:robotwin50_main_results}
\centering
\resizebox{0.95\textwidth}{!}{%
\begin{tabular}{@{}l cc cc cc cc cc cc cc@{}}
\toprule
\multirow{2}{*}{\textbf{Task}} & \multicolumn{2}{c}{Motus} & \multicolumn{2}{c}{$\pi_{0.5}$} & \multicolumn{2}{c}{LingBot-VA} & \multicolumn{2}{c}{LaWAM} & \multicolumn{2}{c}{AHA-WAM} & \multicolumn{2}{c}{Fast-WAM} & \multicolumn{2}{c@{}}{\textbf{w/ TaskWAM}} \\
\cmidrule(lr){2-3}\cmidrule(lr){4-5}\cmidrule(lr){6-7}\cmidrule(lr){8-9}\cmidrule(lr){10-11}\cmidrule(lr){12-13}\cmidrule(l){14-15}
& Clean & Rand. & Clean & Rand. & Clean & Rand. & Clean & Rand. & Clean & Rand. & Clean & Rand. & Clean ($\Delta$) & Rand.($\Delta$) \\
\midrule
Open Laptop & 95 & 91 & 90 & 96 & 92 & 94 & \textbf{100} & \textbf{100} & 98 & 97 & 98 & \textbf{100} & \textbf{100} & \textbf{100} \\
Place Bread Basket & 91 & 94 & 77 & 64 & 97 & 95 & 92 & 85 & 89 & 94 & 91 & 93 & \textbf{98} & \textbf{97} \\
Place Bread Skillet & 86 & 83 & 85 & 66 & \textbf{95} & 90 & 90 & 83 & 91 & 87 & 90 & 93 & 94 & \textbf{98} \\
Place Object Scale & 88 & 85 & 86 & 80 & \textbf{96} & 95 & 95 & 88 & \textbf{96} & 87 & 90 & 97 & \textbf{96} & \textbf{99} \\
$\cdots$ & \multicolumn{14}{c}{$\cdots$} \\
Put Object Cabinet & 88 & 71 & 80 & 79 & 85 & 87 & 90 & 82 & 91 & 89 & \textbf{94} & 89 & 90 & \textbf{90} \\
Rotate QRcode & 89 & 73 & 89 & 87 & \textbf{96} & \textbf{91} & 94 & 89 & 90 & 90 & 93 & 89 & 87 & \textbf{91} \\
Beat Block Hammer & 95 & 88 & 96 & 93 & 96 & \textbf{98} & 90 & 93 & \textbf{100} & 94 & 99 & 97 & \textbf{100} & \textbf{98} \\
\midrule
\rowcolor{taskwamgainrow} 
\textbf{Average (\%)} & 88.66 & 87.02 & 82.74 & 76.76 & 92.90 & 91.50 & 92.52 & 89.48 & \textbf{93.40} & 92.20 & 91.88 & 91.78 & 
92.70 
{\scriptsize\taskwamgain{(+0.82)}} & \textbf{93.06} \scriptsize\taskwamgain{(+1.28)} \\
\bottomrule
\end{tabular}%
}
\vspace{-4mm}
\end{table*}

\subsection{Main Results}
\label{sec:main_results}
\noindent\textbf{Results on StageBench.}
Table~\ref{tab:stagebench_main_results} shows that TaskWAM consistently improves both WAM backbones across all eight task families.
On FastWAM, it raises average success from 40.9\% to 69.4\% (+28.5\%), surpassing the strongest standalone baseline, $\pi_{0.5}$, at 54.5\%.
The gains range from 15--25\% on kitchen tasks and 28--54\% on bimanual tasks, with the largest improvement on Size-Order Cubes (11\% to 65\%).
TaskWAM also improves Cosmos Policy from 32.6\% to 45.5\% (+12.9\%), demonstrating that task-state grounding provides consistent benefits across backbones and different stage-confusion scenarios.

\noindent\textbf{Generalization to Robotwin 2.0.}
Table~\ref{tab:robotwin50_main_results} compares different methods on 50 RoboTwin 2.0 tasks under clean and randomized settings. TaskWAM consistently improves the Fast-WAM backbone and achieves highly competitive performance against recent SOTA methods, while obtaining the best result under the randomized setting. 
These results demonstrate the effectiveness of TaskWAM across diverse manipulation tasks and environmental variations.

\subsection{Ablation and Analysis}
\label{sec:ablation}

\noindent\textbf{Component Ablation.}
Table~\ref{tab:component_ablation} validates the contribution of each TaskWAM component under the \begin{wraptable}[7]{r}{0.4\columnwidth}
\centering
\vspace{-2mm}
\caption{\textbf{Component Ablation.}}
\label{tab:component_ablation}
\scriptsize
\resizebox{\linewidth}{!}{%
\begin{tabular}{@{}lccc|c@{}}
\toprule
\textbf{Method}
& \textbf{Mode}
& \textbf{State}
& \textbf{Residual Flow}
& \textbf{\textit{Suc.}$\uparrow$} \\
\midrule
Baseline
& & & & 40.9 \\
+ Mode Tokens
& \checkmark & & & 56.4 \\
+ State Tokens
& \checkmark & \checkmark & & 61.6\\
\textbf{Full TaskWAM}
& \checkmark & \checkmark & \checkmark  & \textbf{69.4} \\
\bottomrule
\end{tabular}%
}
\end{wraptable} same StageBench setting.
Starting from the 40.9\% baseline, progressively adding Mode Tokens raises \textit{Suc.} to 56.4\%, State Tokens further improve it to 61.6\%, and residual flow steering yields the full-model result of 69.4\%.
These consistent gains confirm that each component contributes to
intended future generation and reliable stage progression.

\begin{wrapfigure}[12]{r}{0.42\linewidth}
    \vspace{-4mm}
    \centering
    \includegraphics[width=\linewidth]{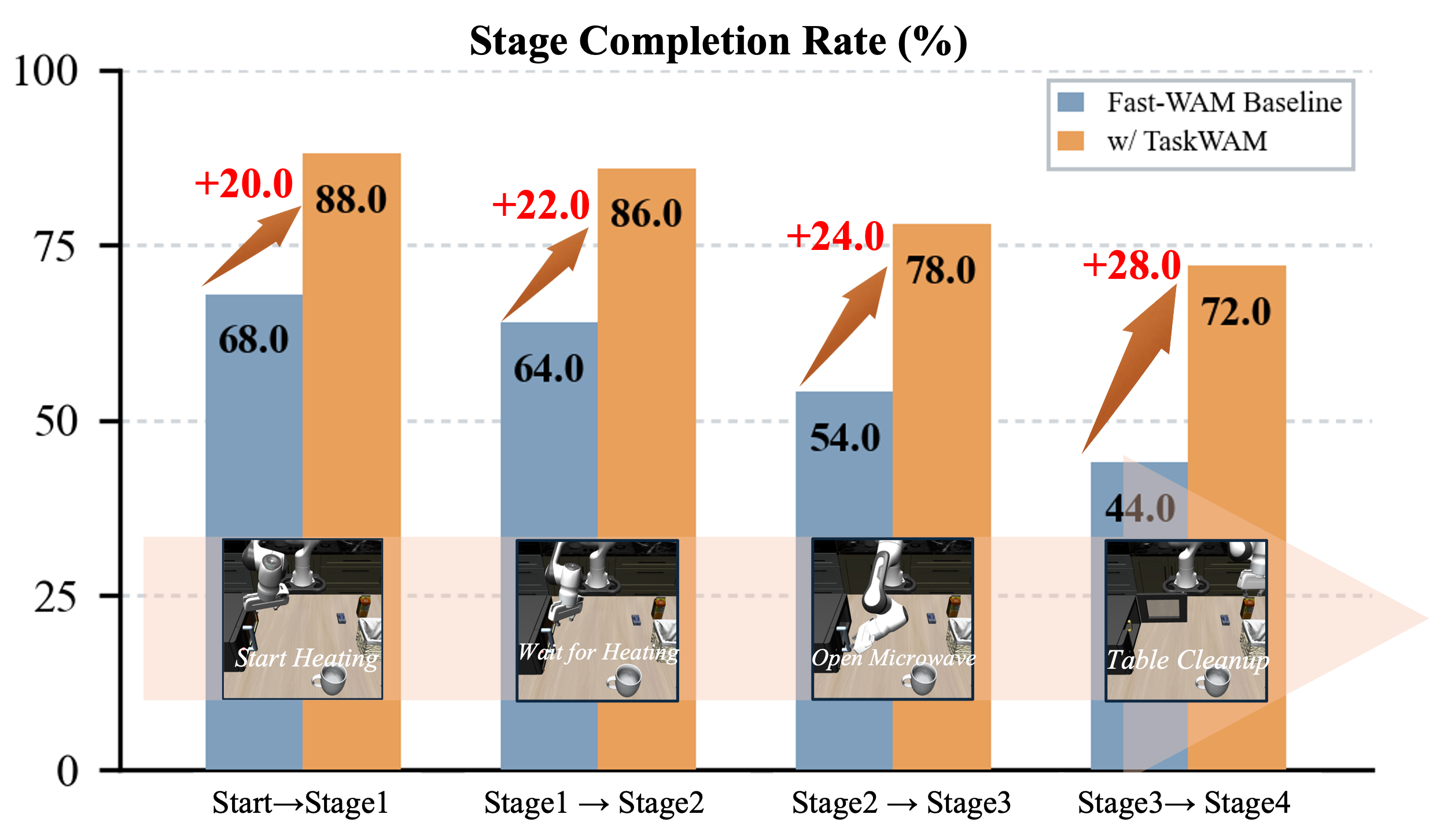}
    \caption{\textbf{Stage-wise completion} over long-horizon execution.
    TaskWAM maintains higher completion across all stage transitions.} 
    \label{fig:stage_completion}
\end{wrapfigure}

\noindent\textbf{Effectiveness of Task-State Grounding.}
Beyond final success rates, we provide complementary benchmark-level and 
transition-level analyses.
More details are provided in Appendix~\ref{app:stage_bench}.
\ding{182} \begin{wraptable}{r}{0.35\columnwidth}
\vspace{-4mm}
\centering
\caption{\textbf{Stage Progression Rate} (\textit{SPR}, \%$\uparrow$) on StageBench.}
\label{tab:spr_analysis}
\scriptsize
\resizebox{\linewidth}{!}{%
\begin{tabular}{l|cc}
\toprule
\textbf{Method}
& \textbf{\textit{Suc.}$\uparrow$}
& \textbf{\textit{SPR}$\uparrow$}
\\
\midrule
FastWAM Baseline
& 40.9 & 68.4 \\

w/ TaskWAM
& 69.4 & 87.1 \\

\bottomrule
\end{tabular}%
}
\vspace{-2mm}
\end{wraptable} \textbf{TaskWAM reduces stage confusion.}
\textit{SPR} measures the fraction of all eligible transitions that complete the correct next stage without filtering failed rollouts; higher is better.
TaskWAM raises \textit{SPR} from 68.4\% to 87.1\%, demonstrating more reliable completion of the intended next stage. 
\ding{183} \textbf{TaskWAM sustains progression across stage transitions.}
In the Cooking \& Cleanup case study (Figure~\ref{fig:stage_completion}), 
TaskWAM achieves higher completion across all transitions, with its advantage widening at later stages, indicating less accumulation of stage confusion over the execution horizon.

Moreover, explicit task-state grounding also improves training
efficiency: TaskWAM converges faster and achieves lower action loss than
FastWAM under the same training setting. 
The convergence curves and full analysis are provided in Appendix~\ref{app:training_convergence} (Figure~\ref{fig:action_convergence}).

Finally, we verify that temporal ordering and execution history are essential for reliable progress tracking.
Additional analysis of State Token design is provided in
Appendix~\ref{app:additional_analysis} (Table~\ref{tab:state_structure}).
\subsection{Real-World Experiments}
\begin{figure*}[t]
    \centering
    \includegraphics[width=0.98\textwidth]{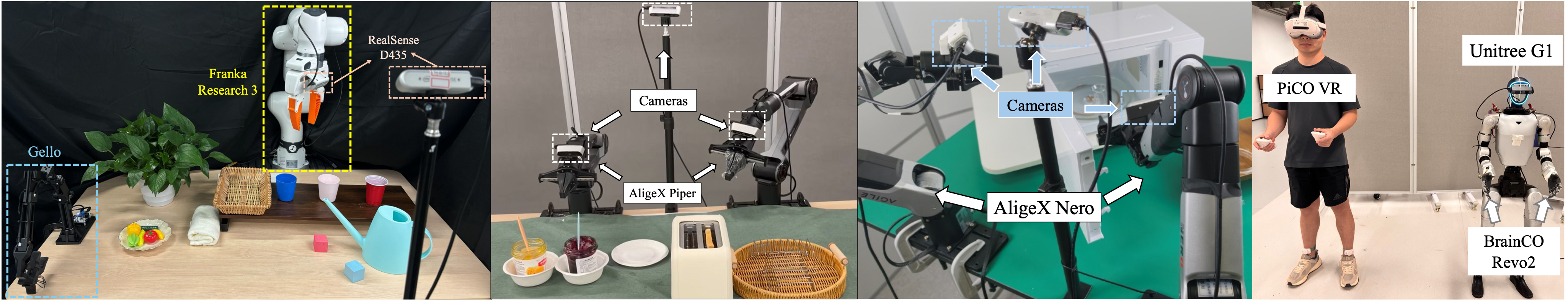}
\caption{\textbf{Real-world robot platform.}
(a) Franka Research 3 with third-person and wrist-view RealSense D435 cameras;
(b-c) dual-arm AgileX Piper (6-DoF) and Nero (7-DoF) equipped with two wrist cameras and one head-mounted camera, respectively;
(c) Unitree G1 humanoid robot equipped with BrainCo Revo2 dexterous hands and operated via PICO VR for whole-body teleoperation.}
    \vspace{-2mm}
    \label{fig:real_world}
\end{figure*}

\noindent\textbf{Robot Platform.}
As shown in Figure~\ref{fig:real_world}, we evaluate our method on gripper-arm, dexterous-hand, and humanoid platforms.
For each task, we collect 100-150 teleoperated demonstrations using Gello~\cite{wu2024gello} for gripper arms, TypeTele~\cite{lin2025typetelereleasingdexterityteleoperation} for dexterous hands, and SONIC~\cite{sonic} for the humanoid platform.
Visual observations are captured by RealSense D435 cameras, providing wrist-mounted and third-person views.

\noindent\textbf{Multi-Stage Real-World Tasks.}
We evaluate our method on four multi-stage tasks drawn from three representative real-world scenarios (bottom in Figure\ref{fig:teaser}):
\ding{182} \textbf{Sequential Cup Nesting}; 
Breakfast Preparation, including \ding{183} \textbf{Bread Toasting \& Spreading} and \ding{184} \textbf{Bread Cutting}; and
\ding{185} \textbf{Laboratory Fluid Manipulation}.
These settings naturally reuse objects and local interactions across tasks and execution stages, providing real-world evaluation of stage confusion under varying task requirements and execution progress.
Each task contains 1-3 variants, with each variant evaluated over 10-20 trials.
Full details are provided in Appendix~\ref{app:real_world}.

\noindent\textbf{Baseline.}
We compare our method with two baseline WAMs, including Cosmos Policy\cite{kim2026cosmospolicy} and Fast-WAM \cite{yuan2026fastwam}.
We report stage-wise completion rates, where $S_k$ denotes cumulative success through stage $k$, with qualitative stage definitions shown in Figure~\ref{fig:teaser} (bottom).

\begin{table}[t]
\caption{\textbf{Results of real-world experiments.}
We report the stage-wise success rate for each task, $S_k$ denotes cumulative success through stage $k$.
Best results are \textbf{bolded}.}
\label{tab:real_world_results}
\centering
\setlength{\tabcolsep}{3.2pt}
\renewcommand{\arraystretch}{1.1}
\resizebox{\columnwidth}{!}{%
\begin{tabular}{l| l| cc cc cc |c}
\toprule
\multirow{2}{*}{\textbf{Task}}
& \multirow{2}{*}{\textbf{Setting}}
& \multicolumn{2}{c}{\textbf{Cosmos Policy}}
& \multicolumn{2}{c}{\textbf{FastWAM}}
& \multicolumn{2}{c}{\textbf{TaskWAM}}
& \multirow{2}{*}{$\boldsymbol{\Delta}$} \\
\cmidrule(lr){3-4}
\cmidrule(lr){5-6}
\cmidrule(l){7-8}
&
& \multicolumn{2}{c}{\textbf{$S_1 \rightarrow S_2 \rightarrow S_3$}}
& \multicolumn{2}{c}{\textbf{$S_1 \rightarrow S_2 \rightarrow S_3$}}
& \multicolumn{2}{c}{\textbf{$S_1 \rightarrow S_2 \rightarrow S_3$}}
& \\
\midrule
Sequential Cup Nesting
& Single-Arm Franka
& \multicolumn{2}{c}{45.0 $\rightarrow$ 40.0 $\rightarrow$ 30.0}
& \multicolumn{2}{c}{55.0 $\rightarrow$ 45.0 $\rightarrow$ 35.0}
& \multicolumn{2}{c}{\textbf{70.0} $\rightarrow$ \textbf{65.0} $\rightarrow$ \textbf{55.0}}
& \taskwamgain{+20.0} \\

Bread Cutting
& Dexterous Hands
& \multicolumn{2}{c}{95.0 $\rightarrow$ 80.0 $\rightarrow$ 20.0}
& \multicolumn{2}{c}{\textbf{100.0} $\rightarrow$ 80.0 $\rightarrow$ 35.0}
& \multicolumn{2}{c}{\textbf{100.0} $\rightarrow$ \textbf{95.0} $\rightarrow$ \textbf{70.0}}
& \taskwamgain{+35.0} \\

Lab. Fluid Manipulation
& Bi-Gripper Nero
& \multicolumn{2}{c}{65.0 $\rightarrow$ 40.0 $\rightarrow$ 40.0}
& \multicolumn{2}{c}{80.0 $\rightarrow$ 55.0 $\rightarrow$ 50.0}
& \multicolumn{2}{c}{\textbf{90.0} $\rightarrow$ \textbf{70.0} $\rightarrow$ \textbf{70.0}}
& \taskwamgain{+20.0} \\
\midrule

\multirow{2}{*}{Bread Toasting \& Spreading}
& \multirow{2}{*}{Bi-Gripper Piper}
& \multicolumn{2}{c}{\textbf{$S_1 \rightarrow S_2 \rightarrow S_3 \rightarrow S_4 \rightarrow S_5$}}
& \multicolumn{2}{c}{\textbf{$S_1 \rightarrow S_2 \rightarrow S_3 \rightarrow S_4 \rightarrow S_5$}}
& \multicolumn{2}{c}{\textbf{$S_1 \rightarrow S_2 \rightarrow S_3 \rightarrow S_4 \rightarrow S_5$}}
& \\
\cmidrule(lr){3-4}
\cmidrule(lr){5-6}
\cmidrule(l){7-8}
&
& \multicolumn{2}{c}{90.0 $\rightarrow$ 20.0 $\rightarrow$ 20.0 $\rightarrow$ 0.0 $\rightarrow$ 0.0}
& \multicolumn{2}{c}{\textbf{95.0} $\rightarrow$ 25.0 $\rightarrow$ 25.0 $\rightarrow$ 10.0 $\rightarrow$ 5.0}
& \multicolumn{2}{c}{\textbf{95.0} $\rightarrow$ \textbf{45.0} $\rightarrow$ \textbf{45.0} $\rightarrow$ \textbf{30.0} $\rightarrow$ \textbf{20.0}}
& \taskwamgain{+15.0} \\

\bottomrule
\end{tabular}%
}
\vspace{-4mm}
\end{table}
\noindent\textbf{Results Analysis.}
Table~\ref{tab:real_world_results} evaluates whether TaskWAM can preserve reliable stage progression in real-world environments.
Across four tasks and diverse embodiments, TaskWAM consistently achieves higher final-stage success than FastWAM, improving by 15--35\%.
The gains become larger in longer tasks with more stage transitions, where stage confusion accumulates over execution.
For instance, in the challenging nearly 2-minute Bread Toasting \& Spreading task, both methods achieve similar early-stage completion, such as toasting. However, after the long waiting stage, FastWAM often confuses the visually similar retrieval state with earlier stages, reducing final-stage success to 5.0\%, while TaskWAM maintains 20.0\% completion.
A representative failure case involving visually similar stages after long waiting periods is further analyzed in Appendix~\ref{app:qualitative}.

\subsection{Results on Heterogeneous Embodiments}
To assess cross-embodiment adaptability, we deploy TaskWAM on four real-world platforms with the same framework while adapting only robot-specific observations and action spaces, as demonstrated in Figure~\ref{fig:teaser} (bottom).
As shown by Table~\ref{tab:real_world_results}, TaskWAM consistently improves final-stage success by 15--35\% over the baseline and exhibits strong adaptability to heterogeneous embodiments, including single-arm robots, 6-DoF and 7-DoF dual-arm platforms, and humanoid with bi-dexterous hands.
Additional details are provided in Appendix~\ref{app:real_world}.

\begin{wraptable}[7]{r}{0.4\columnwidth}
\vspace{-6mm}
\centering
\caption{\textbf{Performance-efficiency trade-off.}
TaskWAM achieves significant gains in multi-stage reliability with
minimal inference overhead.}
\label{tab:efficiency}
\scriptsize
\resizebox{\linewidth}{!}{%
\begin{tabular}{lccc}
\toprule
\textbf{Method}
& \textbf{\textit{Suc.}$\uparrow$}
& \textbf{\textit{SPR.}$\uparrow$} 
& \textbf{Latency (s)$\downarrow$}\\
\midrule
FastWAM      &40.9 &68.4& 0.190\\
\textbf{TaskWAM} & \textbf{69.4} & \textbf{87.1} & 0.209\\
\bottomrule
\end{tabular}%
}
\vspace{-8pt}
\end{wraptable}
\subsection{Performance-Efficiency Analysis}
Table~\ref{tab:efficiency} reports task performance and end-to-end
inference latency under identical hardware settings.
With 30 warm-up and 100 timed runs on a single RTX 5090 GPU, TaskWAM improves \textit{Suc.} from $40.9\%\rightarrow69.4\%$ with only 19 ms additional latency over FastWAM, demonstrating an effective
performance-efficiency trade-off validated by real-world deployment.
\section{conclusion}
Reliable stage progression is a critical capability for long-horizon robotic manipulation,  especially when different tasks and execution stages share similar observation-action interactions.
In this work, we identify \emph{stage confusion} in World-Action Models, where locally plausible yet misdirected rollouts to drift toward wrong task stage, and propose TaskWAM to address this problem.
Our key insight is to ground future generation in explicit task requirements and realized execution progress through mode codebook retrieval, causal execution tracking, and backbone-agnostic future steering.
To support systematic evaluation, we further introduce StageBench with overlapping multi-stage trajectories and stage-aware evaluation.
Extensive experiments in simulation and real-world settings validate TaskWAM across heterogeneous WAM architectures and diverse robot embodiments.
In summary, our results highlight explicit stage-aware future generation as an important step toward reliable generalist robot behavior in long-horizon tasks.

\subsection*{AI use statement}
In this work, we used generative AI tools for checking the consistency of
technical writing, and identifying possible typographical or formatting errors in the manuscript.
We have not used generative AI tools for developing the research methodology, and tasks involving datasets, experiments, implementation, qualitative data analysis,
and the interpretation of experimental results are not applicable to this work.
Additionally, we used generative AI tools for improving the grammar and readability of the manuscript. We have reviewed all AI-assisted work. We reviewed all AI-assisted revisions and manually verified
that they preserved the intended technical meaning and did not alter the
methodology, experimental results, or conclusions. We take responsibility for the final content of this work,
including text, claims or artifacts produced with the aid of generative AI.





\subsection*{Reproducibility statement}
We provide the implementation of our method and detailed experimental
configurations in the supplementary material. We include model architectures,
training settings, hyperparameters, benchmark construction details, and
evaluation protocols to facilitate faithful reproduction of our results.




\bibliography{iclr2027_conference}

\begin{thebibliography}{38}
\providecommand{\natexlab}[1]{#1}
\providecommand{\url}[1]{\texttt{#1}}
\expandafter\ifx\csname urlstyle\endcsname\relax
  \providecommand{\doi}[1]{doi: #1}\else
  \providecommand{\doi}{doi: \begingroup \urlstyle{rm}\Url}\fi

\bibitem[Belkhale et~al.(2024)Belkhale, Ding, Xiao, Sermanet, Vuong, Tompson,
  Chebotar, Dwibedi, and Sadigh]{Belkhale-RSS-24}
Suneel Belkhale, Tianli Ding, Ted Xiao, Pierre Sermanet, Quan Vuong, Jonathan
  Tompson, Yevgen Chebotar, Debidatta Dwibedi, and Dorsa Sadigh.
\newblock {RT-H: Action Hierarchies using Language}.
\newblock In \emph{Proceedings of Robotics: Science and Systems}, Delft,
  Netherlands, July 2024.
\newblock \doi{10.15607/RSS.2024.XX.049}.

\bibitem[Bi et~al.(2026)Bi, Tan, Xie, Wang, Huang, Liu, Zhao, Feng, Xiang,
  Rong, Zhao, Liu, Su, Ma, Su, and Zhu]{bi2026motus}
Hongzhe Bi, Hengkai Tan, Shenghao Xie, Zeyuan Wang, Shuhe Huang, Haitian Liu,
  Ruowen Zhao, Yao Feng, Chendong Xiang, Yinze Rong, Hongyan Zhao, Hanyu Liu,
  Zhizhong Su, Lei Ma, Hang Su, and Jun Zhu.
\newblock Motus: A unified latent action world model.
\newblock In \emph{Proceedings of the IEEE/CVF Conference on Computer Vision
  and Pattern Recognition (CVPR)}, pp.\  35101--35113, June 2026.

\bibitem[Black et~al.(2025{\natexlab{a}})Black, Brown, Darpinian, Dhabalia,
  Driess, Esmail, Equi, Finn, Fusai, Galliker, Ghosh, Groom, Hausman, ichter,
  Jakubczak, Jones, Ke, LeBlanc, Levine, Li-Bell, Mothukuri, Nair, Pertsch,
  Ren, Shi, Smith, Springenberg, Stachowicz, Tanner, Vuong, Walke, Walling,
  Wang, Yu, and Zhilinsky]{physicalintelligence2025pi05}
Kevin Black, Noah Brown, James Darpinian, Karan Dhabalia, Danny Driess, Adnan
  Esmail, Michael~Robert Equi, Chelsea Finn, Niccolo Fusai, Manuel~Y. Galliker,
  Dibya Ghosh, Lachy Groom, Karol Hausman, brian ichter, Szymon Jakubczak, Tim
  Jones, Liyiming Ke, Devin LeBlanc, Sergey Levine, Adrian Li-Bell, Mohith
  Mothukuri, Suraj Nair, Karl Pertsch, Allen~Z. Ren, Lucy~Xiaoyang Shi, Laura
  Smith, Jost~Tobias Springenberg, Kyle Stachowicz, James Tanner, Quan Vuong,
  Homer Walke, Anna Walling, Haohuan Wang, Lili Yu, and Ury Zhilinsky.
\newblock $\pi_{0.5}$: a vision-language-action model with open-world
  generalization.
\newblock In Joseph Lim, Shuran Song, and Hae-Won Park (eds.),
  \emph{Proceedings of The 9th Conference on Robot Learning}, volume 305 of
  \emph{Proceedings of Machine Learning Research}, pp.\  17--40. PMLR, 27--30
  Sep 2025{\natexlab{a}}.

\bibitem[Black et~al.(2025{\natexlab{b}})Black, Galliker, and
  Levine]{NEURIPS2025_300ccb21}
Kevin Black, Manuel Galliker, and Sergey Levine.
\newblock Real-time execution of action chunking flow policies.
\newblock In \emph{Advances in Neural Information Processing Systems}, volume
  38, Main Conference, pp.\  33383--33407. Curran Associates, Inc.,
  2025{\natexlab{b}}.
\newblock \doi{10.52202/085713-1122}.

\bibitem[Cai et~al.(2026)Cai, Ling, Chu, Liu, Kang, Liang, Xu, Mao, Zhang,
  Yang, Ying, Zheng, and Mu]{cai2026ahawam}
Jisong Cai, Long Ling, Shiwei Chu, Zhongshan Liu, Jiayue Kang, Zhixuan Liang,
  Wenjie Xu, Yinan Mao, Weinan Zhang, Xiaokang Yang, Ru~Ying, Ran Zheng, and
  Yao Mu.
\newblock Aha-wam:asynchronous horizon-adaptive world-action modeling with
  observation-guided context routing, 2026.
\newblock URL \url{https://arxiv.org/abs/2606.09811}.

\bibitem[Chen et~al.(2026{\natexlab{a}})Chen, Wang, Chen, Chen, Gao, Tang, Li,
  Liu, Yao, Li, Xu, and Yu]{chen2026lawam}
Jialei Chen, Kai Wang, Kang Chen, Shuaihang Chen, Feng Gao, Wenhao Tang,
  Zhiyuan Li, Weilin Liu, Zhuyu Yao, Boxun Li, Yuanbo Xu, and Chao Yu.
\newblock Lawam: Latent world action models for efficient dynamics-aware robot
  policies, 2026{\natexlab{a}}.
\newblock URL \url{https://arxiv.org/abs/2606.15768}.

\bibitem[Chen et~al.(2026{\natexlab{b}})Chen, Chen, Chen, Cai, Liu, Li, Liang,
  Lin, Ge, Gu, Deng, Guo, Nian, Xie, Chen, Su, Xu, Liu, Hu, ang Gao, Wang,
  Liang, Qin, Yang, Luo, and Mu]{chen2025robotwin2}
Tianxing Chen, Zanxin Chen, Baijun Chen, Zijian Cai, Yibin Liu, Zixuan Li,
  Qiwei Liang, Xianliang Lin, Yiheng Ge, Zhenyu Gu, Weiliang Deng, Yubin Guo,
  Tian Nian, Xuanbing Xie, Qiangyu Chen, Kailun Su, Tianling Xu, Guodong Liu,
  Mengkang Hu, Huan ang Gao, Kaixuan Wang, Zhixuan Liang, Yusen Qin, Xiaokang
  Yang, Ping Luo, and Yao Mu.
\newblock Robotwin 2.0: A scalable data generator and benchmark with strong
  domain randomization for robust bimanual robotic manipulation.
\newblock In \emph{Forty-third International Conference on Machine Learning},
  2026{\natexlab{b}}.

\bibitem[Chung et~al.(2026)Chung, Hanyu, Nguyen, Le, Bumgarner, Nguyen, Vo,
  Yamazaki, Rainwater, Kieu, Nguyen, and Le]{chung2026memorystate}
Nhat Chung, Taisei Hanyu, Toan Nguyen, Huy Le, Frederick Bumgarner, Duy Minh~Ho
  Nguyen, Khoa Vo, Kashu Yamazaki, Chase Rainwater, Tung Kieu, Anh Nguyen, and
  Ngan Le.
\newblock Rethinking progression of memory state in robotic manipulation: An
  object-centric perspective.
\newblock \emph{Proceedings of the AAAI Conference on Artificial Intelligence},
  40\penalty0 (5):\penalty0 3407--3415, 2026.
\newblock \doi{10.1609/aaai.v40i5.37337}.

\bibitem[Dai et~al.(2019)Dai, Yang, Yang, Carbonell, Le, and
  Salakhutdinov]{dai-etal-2019-transformer}
Zihang Dai, Zhilin Yang, Yiming Yang, Jaime Carbonell, Quoc Le, and Ruslan
  Salakhutdinov.
\newblock Transformer-{XL}: Attentive language models beyond a fixed-length
  context.
\newblock In Anna Korhonen, David Traum, and Llu{\'i}s M{\`a}rquez (eds.),
  \emph{Proceedings of the 57th Annual Meeting of the Association for
  Computational Linguistics}, pp.\  2978--2988, Florence, Italy, July 2019.
  Association for Computational Linguistics.
\newblock \doi{10.18653/v1/P19-1285}.

\bibitem[Fang et~al.(2025)Fang, Grotz, Pumacay, Wang, Fox, Krishna, and
  Duan]{pmlr-v267-fang25c}
Haoquan Fang, Markus Grotz, Wilbert Pumacay, Yi~Ru Wang, Dieter Fox, Ranjay
  Krishna, and Jiafei Duan.
\newblock {SAM}2{A}ct: Integrating visual foundation model with a memory
  architecture for robotic manipulation.
\newblock In Aarti Singh, Maryam Fazel, Daniel Hsu, Simon Lacoste-Julien, Felix
  Berkenkamp, Tegan Maharaj, Kiri Wagstaff, and Jerry Zhu (eds.),
  \emph{Proceedings of the 42nd International Conference on Machine Learning},
  volume 267 of \emph{Proceedings of Machine Learning Research}, pp.\
  15925--15942. PMLR, 13--19 Jul 2025.

\bibitem[Fang et~al.(2019)Fang, Toshev, Fei-Fei, and Savarese]{Fang_2019_CVPR}
Kuan Fang, Alexander Toshev, Li~Fei-Fei, and Silvio Savarese.
\newblock Scene memory transformer for embodied agents in long-horizon tasks.
\newblock In \emph{Proceedings of the IEEE/CVF Conference on Computer Vision
  and Pattern Recognition (CVPR)}, June 2019.

\bibitem[Guhur et~al.(2023)Guhur, Chen, Pinel, Tapaswi, Laptev, and
  Schmid]{pmlr-v205-guhur23a}
Pierre-Louis Guhur, Shizhe Chen, Ricardo~Garcia Pinel, Makarand Tapaswi, Ivan
  Laptev, and Cordelia Schmid.
\newblock Instruction-driven history-aware policies for robotic manipulations.
\newblock In Karen Liu, Dana Kulic, and Jeff Ichnowski (eds.),
  \emph{Proceedings of The 6th Conference on Robot Learning}, volume 205 of
  \emph{Proceedings of Machine Learning Research}, pp.\  175--187. PMLR, 14--18
  Dec 2023.

\bibitem[Henschel et~al.(2025)Henschel, Khachatryan, Poghosyan, Hayrapetyan,
  Tadevosyan, Wang, Navasardyan, and Shi]{Henschel_2025_CVPR}
Roberto Henschel, Levon Khachatryan, Hayk Poghosyan, Daniil Hayrapetyan, Vahram
  Tadevosyan, Zhangyang Wang, Shant Navasardyan, and Humphrey Shi.
\newblock {StreamingT2V}: Consistent, dynamic, and extendable long video
  generation from text.
\newblock In \emph{Proceedings of the IEEE/CVF Conference on Computer Vision
  and Pattern Recognition (CVPR)}, pp.\  2568--2577, June 2025.

\bibitem[Kim et~al.(2026)Kim, Gao, Lin, Lin, Ge, Lam, Liang, Song, Liu, Finn,
  and Gu]{kim2026cosmospolicy}
Moo~Jin Kim, Yihuai Gao, Tsung-Yi Lin, Yen-Chen Lin, Yunhao Ge, Grace Lam,
  Percy Liang, Shuran Song, Ming-Yu Liu, Chelsea Finn, and Jinwei Gu.
\newblock Cosmos policy: Fine-tuning video models for visuomotor control and
  planning.
\newblock In \emph{International Conference on Learning Representations},
  volume 2026, pp.\  71531--71552, 2026.

\bibitem[Li et~al.(2023{\natexlab{a}})Li, Zhang, Wong, Gokmen, Srivastava,
  Mart\'in-Mart\'in, Wang, Levine, Lingelbach, Sun, Anvari, Hwang, Sharma,
  Aydin, Bansal, Hunter, Kim, Lou, Matthews, Villa-Renteria, Tang, Tang, Xia,
  Savarese, Gweon, Liu, Wu, and Fei-Fei]{pmlr-v205-li23a}
Chengshu Li, Ruohan Zhang, Josiah Wong, Cem Gokmen, Sanjana Srivastava, Roberto
  Mart\'in-Mart\'in, Chen Wang, Gabrael Levine, Michael Lingelbach, Jiankai
  Sun, Mona Anvari, Minjune Hwang, Manasi Sharma, Arman Aydin, Dhruva Bansal,
  Samuel Hunter, Kyu-Young Kim, Alan Lou, Caleb~R Matthews, Ivan
  Villa-Renteria, Jerry~Huayang Tang, Claire Tang, Fei Xia, Silvio Savarese,
  Hyowon Gweon, Karen Liu, Jiajun Wu, and Li~Fei-Fei.
\newblock Behavior-1k: A benchmark for embodied ai with 1,000 everyday
  activities and realistic simulation.
\newblock In Karen Liu, Dana Kulic, and Jeff Ichnowski (eds.),
  \emph{Proceedings of The 6th Conference on Robot Learning}, volume 205 of
  \emph{Proceedings of Machine Learning Research}, pp.\  80--93. PMLR, 14--18
  Dec 2023{\natexlab{a}}.

\bibitem[Li et~al.(2023{\natexlab{b}})Li, Li, Savarese, and
  Hoi]{pmlr-v202-li23q}
Junnan Li, Dongxu Li, Silvio Savarese, and Steven Hoi.
\newblock {BLIP}-2: Bootstrapping language-image pre-training with frozen image
  encoders and large language models.
\newblock In Andreas Krause, Emma Brunskill, Kyunghyun Cho, Barbara Engelhardt,
  Sivan Sabato, and Jonathan Scarlett (eds.), \emph{Proceedings of the 40th
  International Conference on Machine Learning}, volume 202 of
  \emph{Proceedings of Machine Learning Research}, pp.\  19730--19742. PMLR,
  23--29 Jul 2023{\natexlab{b}}.

\bibitem[Li et~al.(2026)Li, Zhang, Luo, Yang, Wang, Zhang, Yu, Gao, Xue, Zhou,
  Zhu, Ding, Shen, and Xu]{li2026lingbotva}
Lin Li, Qihang Zhang, Yiming Luo, Shuai Yang, Ruilin Wang, Luyao Zhang, Mingrui
  Yu, Zelin Gao, Nan Xue, Boyu Zhou, Xing Zhu, Mingyu Ding, Yujun Shen, and
  Yinghao Xu.
\newblock {Causal World Modeling for Robot Control}.
\newblock In \emph{Proceedings of Robotics: Science and Systems}, Sydney,
  Australia, July 2026.
\newblock \doi{10.15607/RSS.2026.XXII.016}.

\bibitem[Li et~al.(2025)Li, Gao, Sadigh, and Song]{li2025uva}
Shuang Li, Yihuai Gao, Dorsa Sadigh, and Shuran Song.
\newblock {Unified Video Action Model}.
\newblock In \emph{Proceedings of Robotics: Science and Systems}, LosAngeles,
  CA, USA, June 2025.
\newblock \doi{10.15607/RSS.2025.XXI.074}.

\bibitem[Lin et~al.(2026)Lin, He, Bao, Zhao, Li, Wang, Wang, Chi, and
  Zhang]{lin2026jepawamlearningvisionlanguageactionpolicies}
Yihan Lin, Jiawei He, Shifeng Bao, Chen Zhao, Yang Li, Xiaobo Wang, Yan Wang,
  Cheng Chi, and Jing Zhang.
\newblock {JEPA-WAM}: Learning vision-language-action policies with
  joint-embedding world modeling, 2026.

\bibitem[Lin et~al.(2025)Lin, Wei, Liao, Lin, Xing, Li, Zhang, Cutkosky, and
  Zheng]{lin2025typetelereleasingdexterityteleoperation}
Yuhao Lin, Yi-Lin Wei, Haoran Liao, Mu~Lin, Chengyi Xing, Hao Li, Dandan Zhang,
  Mark Cutkosky, and Wei-Shi Zheng.
\newblock Typetele: Releasing dexterity in teleoperation by dexterous
  manipulation types.
\newblock In Joseph Lim, Shuran Song, and Hae-Won Park (eds.),
  \emph{Proceedings of The 9th Conference on Robot Learning}, volume 305 of
  \emph{Proceedings of Machine Learning Research}, pp.\  4975--4993. PMLR,
  27--30 Sep 2025.

\bibitem[Liu et~al.(2023)Liu, Zhu, Gao, Feng, Liu, Zhu, and
  Stone]{NEURIPS2023_8c3c6668}
Bo~Liu, Yifeng Zhu, Chongkai Gao, Yihao Feng, Qiang Liu, Yuke Zhu, and Peter
  Stone.
\newblock Libero: Benchmarking knowledge transfer for lifelong robot learning.
\newblock In A.~Oh, T.~Naumann, A.~Globerson, K.~Saenko, M.~Hardt, and
  S.~Levine (eds.), \emph{Advances in Neural Information Processing Systems},
  volume~36, pp.\  44776--44791. Curran Associates, Inc., 2023.
\newblock \doi{10.52202/075280-1939}.

\bibitem[Liu et~al.(2025)Liu, Wu, Li, Tan, Chen, Wang, Xu, Su, and
  Zhu]{ICLR2025_49f80e4d}
Songming Liu, Lingxuan Wu, Bangguo Li, Hengkai Tan, Huayu Chen, Zhengyi Wang,
  Ke~Xu, Hang Su, and Jun Zhu.
\newblock Rdt-1b: a diffusion foundation model for bimanual manipulation.
\newblock In Y.~Yue, A.~Garg, N.~Peng, F.~Sha, and R.~Yu (eds.),
  \emph{International Conference on Learning Representations}, volume 2025,
  pp.\  29982--30009, 2025.

\bibitem[Luo et~al.(2026)Luo, Yuan, Wang, Li, Casta{\~{n}}eda, Chen, Cao, Li,
  Minor, Ben, Park, Sami, Wang, Da, Ding, Hogg, Song, Lim, Jeong, He, Xue,
  Xiao, Yuen, Kautz, Chang, Iqbal, Fan, and Zhu]{sonic}
Zhengyi Luo, Ye~Yuan, Tingwu Wang, Chenran Li, Fernando Casta{\~{n}}eda, Sirui
  Chen, Zi{-}Ang Cao, Jiefeng Li, David Minor, Qingwei Ben, Jinhyung Park,
  David Sami, Zi~Wang, Xingye Da, Runyu Ding, Cyrus Hogg, Lina Song, Edy Lim,
  Eugene Jeong, Tairan He, Haoru Xue, Wenli Xiao, Simon Yuen, Jan Kautz, Yan
  Chang, Umar Iqbal, Linxi~Jim Fan, and Yuke Zhu.
\newblock {SONIC:} supersizing motion tracking for natural humanoid whole-body
  control.
\newblock \emph{Sci. Robotics}, 11\penalty0 (117):\penalty0 eaed4592, 2026.
\newblock \doi{10.1126/SCIROBOTICS.AED4592}.

\bibitem[Mees et~al.(2022)Mees, Hermann, Rosete-Beas, and
  Burgard]{mees2022calvin}
Oier Mees, Lukas Hermann, Erick Rosete-Beas, and Wolfram Burgard.
\newblock Calvin: A benchmark for language-conditioned policy learning for
  long-horizon robot manipulation tasks.
\newblock \emph{IEEE Robotics and Automation Letters}, 7\penalty0 (3):\penalty0
  7327--7334, 2022.
\newblock \doi{10.1109/LRA.2022.3180108}.

\bibitem[Nasiriany et~al.(2026)Nasiriany, Nasiriany, Maddukuri, and
  Zhu]{ICLR2026_a05003fd}
Soroush Nasiriany, Sep Nasiriany, Abhiram Maddukuri, and Yuke Zhu.
\newblock Robocasa365: A large-scale simulation framework for training and
  benchmarking generalist robots.
\newblock In \emph{International Conference on Learning Representations},
  volume 2026, pp.\  98643--98667, 2026.

\bibitem[NVIDIA et~al.(2025)NVIDIA, :, Bjorck, Castañeda, Cherniadev, Da,
  Ding, Fan, Fang, Fox, Hu, Huang, Jang, Jiang, Kautz, Kundalia, Lao, Li, Lin,
  Lin, Liu, Llontop, Magne, Mandlekar, Narayan, Nasiriany, Reed, Tan, Wang,
  Wang, Wang, Wang, Xiang, Xie, Xu, Xu, Ye, Yu, Zhang, Zhang, Zhao, Zheng, and
  Zhu]{nvidia2026gr00tn17}
NVIDIA, :, Johan Bjorck, Fernando Castañeda, Nikita Cherniadev, Xingye Da,
  Runyu Ding, Linxi~"Jim" Fan, Yu~Fang, Dieter Fox, Fengyuan Hu, Spencer Huang,
  Joel Jang, Zhenyu Jiang, Jan Kautz, Kaushil Kundalia, Lawrence Lao, Zhiqi Li,
  Zongyu Lin, Kevin Lin, Guilin Liu, Edith Llontop, Loic Magne, Ajay Mandlekar,
  Avnish Narayan, Soroush Nasiriany, Scott Reed, You~Liang Tan, Guanzhi Wang,
  Zu~Wang, Jing Wang, Qi~Wang, Jiannan Xiang, Yuqi Xie, Yinzhen Xu, Zhenjia Xu,
  Seonghyeon Ye, Zhiding Yu, Ao~Zhang, Hao Zhang, Yizhou Zhao, Ruijie Zheng,
  and Yuke Zhu.
\newblock Gr00t n1: An open foundation model for generalist humanoid robots,
  2025.
\newblock URL \url{https://arxiv.org/abs/2503.14734}.

\bibitem[O'Neill et~al.(2024)O'Neill, Rehman, Maddukuri, Gupta, Padalkar, Lee,
  Pooley, Gupta, Mandlekar, Jain, Tung, Bewley, Herzog, Irpan, Khazatsky, Rai,
  Gupta, Wang, Singh, Garg, Kembhavi, Xie, Brohan, Raffin, Sharma, Yavary,
  Jain, Balakrishna, Wahid, Burgess{-}Limerick, Kim, Sch{\"{o}}lkopf, Wulfe,
  Ichter, Lu, Xu, Le, Finn, Wang, Xu, Chi, Huang, Chan, Agia, Pan, Fu, Devin,
  Xu, Morton, Driess, Chen, Pathak, Shah, B{\"{u}}chler, Jayaraman,
  Kalashnikov, Sadigh, Johns, Foster, Liu, Ceola, Xia, Zhao, Stulp, Zhou,
  Sukhatme, Salhotra, Yan, Feng, Schiavi, Berseth, Kahn, Wang, Su, Fang, Shi,
  Bao, Amor, Christensen, Furuta, Walke, Fang, Ha, Mordatch, Radosavovic, Leal,
  Liang, Abou{-}Chakra, Kim, Drake, Peters, Schneider, Hsu, Bohg, Bingham, Wu,
  Gao, Hu, Wu, Wu, Sun, Luo, Gu, Tan, Oh, Wu, Lu, Yang, Malik, Silv{\'{e}}rio,
  Hejna, Booher, Tompson, Yang, Salvador, Lim, Han, Wang, Rao, Pertsch,
  Hausman, Go, Gopalakrishnan, Goldberg, Byrne, Oslund, Kawaharazuka, Black,
  Lin, Zhang, Ehsani, Lekkala, Ellis, Rana, Srinivasan, Fang, Singh, Zeng,
  Hatch, Hsu, Itti, Chen, Pinto, Fei{-}Fei, Tan, Fan, Ott, Lee, Weihs, Chen,
  Lepert, Memmel, Tomizuka, Itkina, Castro, Spero, Du, Ahn, Yip, Zhang, Ding,
  Heo, Srirama, Sharma, Kim, Kanazawa, Hansen, Heess, Joshi, S{\"{u}}nderhauf,
  Liu, Palo, Shafiullah, Mees, Kroemer, Bastani, Sanketi, Miller, Yin,
  Wohlhart, Xu, Fagan, Mitrano, Sermanet, Abbeel, Sundaresan, Chen, Vuong,
  Rafailov, Tian, Doshi, Mart{\'{\i}}n{-}Mart{\'{\i}}n, Baijal, Scalise,
  Hendrix, Lin, Qian, Zhang, Mendonca, Shah, Hoque, Julian, Bustamante{-}Gomez,
  Kirmani, Levine, Lin, Moore, Bahl, Dass, Sonawani, Song, Xu, Haldar,
  Karamcheti, Adebola, Guist, Nasiriany, Schaal, Welker, Tian, Ramamoorthy,
  Dasari, Belkhale, Park, Nair, Mirchandani, Osa, Gupta, Harada, Matsushima,
  Xiao, Kollar, Yu, Ding, Davchev, Zhao, Armstrong, Darrell, Chung, Jain,
  Vanhoucke, Zhan, Zhou, Burgard, Chen, Wang, Zhu, Geng, Liu, Xu, Li, Lu, Ma,
  Kim, Chebotar, Zhou, Zhu, Wu, Xu, Wang, Bisk, Cho, Lee, Cui, Cao, Wu, Tang,
  Zhu, Zhang, Jiang, Li, Li, Iwasawa, Matsuo, Ma, Xu, Cui, Zhang, and
  Lin]{OpenX24}
Abby O'Neill, Abdul Rehman, Abhiram Maddukuri, Abhishek Gupta, Abhishek
  Padalkar, Abraham Lee, Acorn Pooley, Agrim Gupta, Ajay Mandlekar, Ajinkya
  Jain, Albert Tung, Alex Bewley, Alexander Herzog, Alex Irpan, Alexander
  Khazatsky, Anant Rai, Anchit Gupta, Andrew~E. Wang, Anikait Singh, Animesh
  Garg, Aniruddha Kembhavi, Annie Xie, Anthony Brohan, Antonin Raffin, Archit
  Sharma, Arefeh Yavary, Arhan Jain, Ashwin Balakrishna, Ayzaan Wahid, Ben
  Burgess{-}Limerick, Beomjoon Kim, Bernhard Sch{\"{o}}lkopf, Blake Wulfe,
  Brian Ichter, Cewu Lu, Charles Xu, Charlotte Le, Chelsea Finn, Chen Wang,
  Chenfeng Xu, Cheng Chi, Chenguang Huang, Christine Chan, Christopher Agia,
  Chuer Pan, Chuyuan Fu, Coline Devin, Danfei Xu, Daniel Morton, Danny Driess,
  Daphne Chen, Deepak Pathak, Dhruv Shah, Dieter B{\"{u}}chler, Dinesh
  Jayaraman, Dmitry Kalashnikov, Dorsa Sadigh, Edward Johns, Ethan~Paul Foster,
  Fangchen Liu, Federico Ceola, Fei Xia, Feiyu Zhao, Freek Stulp, Gaoyue Zhou,
  Gaurav~S. Sukhatme, Gautam Salhotra, Ge~Yan, Gilbert Feng, Giulio Schiavi,
  Glen Berseth, Gregory Kahn, Guanzhi Wang, Hao Su, Haoshu Fang, Haochen Shi,
  Henghui Bao, Heni~Ben Amor, Henrik~I. Christensen, Hiroki Furuta, Homer
  Walke, Hongjie Fang, Huy Ha, Igor Mordatch, Ilija Radosavovic, Isabel Leal,
  Jacky Liang, Jad Abou{-}Chakra, Jaehyung Kim, Jaimyn Drake, Jan Peters, Jan
  Schneider, Jasmine Hsu, Jeannette Bohg, Jeffrey~T. Bingham, Jeffrey Wu,
  Jensen Gao, Jiaheng Hu, Jiajun Wu, Jialin Wu, Jiankai Sun, Jianlan Luo,
  Jiayuan Gu, Jie Tan, Jihoon Oh, Jimmy Wu, Jingpei Lu, Jingyun Yang, Jitendra
  Malik, Jo{\~{a}}o Silv{\'{e}}rio, Joey Hejna, Jonathan Booher, Jonathan
  Tompson, Jonathan Yang, Jordi Salvador, Joseph~J. Lim, Junhyek Han, Kaiyuan
  Wang, Kanishka Rao, Karl Pertsch, Karol Hausman, Keegan Go, Keerthana
  Gopalakrishnan, Ken Goldberg, Kendra Byrne, Kenneth Oslund, Kento
  Kawaharazuka, Kevin Black, Kevin Lin, Kevin Zhang, Kiana Ehsani, Kiran
  Lekkala, Kirsty Ellis, Krishan Rana, Krishnan Srinivasan, Kuan Fang,
  Kunal~Pratap Singh, Kuo{-}Hao Zeng, Kyle Hatch, Kyle Hsu, Laurent Itti,
  Lawrence~Yunliang Chen, Lerrel Pinto, Li~Fei{-}Fei, Liam Tan, Linxi~Jim Fan,
  Lionel Ott, Lisa Lee, Luca Weihs, Magnum Chen, Marion Lepert, Marius Memmel,
  Masayoshi Tomizuka, Masha Itkina, Mateo~Guaman Castro, Max Spero, Maximilian
  Du, Michael Ahn, Michael~C. Yip, Mingtong Zhang, Mingyu Ding, Minho Heo,
  Mohan~Kumar Srirama, Mohit Sharma, Moo~Jin Kim, Naoaki Kanazawa, Nicklas
  Hansen, Nicolas Heess, Nikhil~J. Joshi, Niko S{\"{u}}nderhauf, Ning Liu,
  Norman~Di Palo, Nur Muhammad~(Mahi) Shafiullah, Oier Mees, Oliver Kroemer,
  Osbert Bastani, Pannag~R. Sanketi, Patrick~Tree Miller, Patrick Yin, Paul
  Wohlhart, Peng Xu, Peter~David Fagan, Peter Mitrano, Pierre Sermanet, Pieter
  Abbeel, Priya Sundaresan, Qiuyu Chen, Quan Vuong, Rafael Rafailov, Ran Tian,
  Ria Doshi, Roberto Mart{\'{\i}}n{-}Mart{\'{\i}}n, Rohan Baijal, Rosario
  Scalise, Rose Hendrix, Roy Lin, Runjia Qian, Ruohan Zhang, Russell Mendonca,
  Rutav Shah, Ryan Hoque, Ryan Julian, Samuel Bustamante{-}Gomez, Sean Kirmani,
  Sergey Levine, Shan Lin, Sherry Moore, Shikhar Bahl, Shivin Dass, Shubham~D.
  Sonawani, Shuran Song, Sichun Xu, Siddhant Haldar, Siddharth Karamcheti,
  Simeon Adebola, Simon Guist, Soroush Nasiriany, Stefan Schaal, Stefan Welker,
  Stephen Tian, Subramanian Ramamoorthy, Sudeep Dasari, Suneel Belkhale,
  Sungjae Park, Suraj Nair, Suvir Mirchandani, Takayuki Osa, Tanmay Gupta,
  Tatsuya Harada, Tatsuya Matsushima, Ted Xiao, Thomas Kollar, Tianhe Yu,
  Tianli Ding, Todor Davchev, Tony~Z. Zhao, Travis Armstrong, Trevor Darrell,
  Trinity Chung, Vidhi Jain, Vincent Vanhoucke, Wei Zhan, Wenxuan Zhou, Wolfram
  Burgard, Xi~Chen, Xiaolong Wang, Xinghao Zhu, Xinyang Geng, Xiyuan Liu,
  Liangwei Xu, Xuanlin Li, Yao Lu, Yecheng~Jason Ma, Yejin Kim, Yevgen
  Chebotar, Yifan Zhou, Yifeng Zhu, Yilin Wu, Ying Xu, Yixuan Wang, Yonatan
  Bisk, Yoonyoung Cho, Youngwoon Lee, Yuchen Cui, Yue Cao, Yueh{-}Hua Wu, Yujin
  Tang, Yuke Zhu, Yunchu Zhang, Yunfan Jiang, Yunshuang Li, Yunzhu Li, Yusuke
  Iwasawa, Yutaka Matsuo, Zehan Ma, Zhuo Xu, Zichen~Jeff Cui, Zichen Zhang, and
  Zipeng Lin.
\newblock Open x-embodiment: Robotic learning datasets and {RT-X} models : Open
  x-embodiment collaboration.
\newblock In \emph{{IEEE} International Conference on Robotics and Automation,
  {ICRA} 2024, Yokohama, Japan, May 13-17, 2024}, pp.\  6892--6903. {IEEE},
  2024.
\newblock \doi{10.1109/ICRA57147.2024.10611477}.

\bibitem[Peebles \& Xie(2023)Peebles and Xie]{Peebles_2023_ICCV}
William Peebles and Saining Xie.
\newblock Scalable diffusion models with transformers.
\newblock In \emph{Proceedings of the IEEE/CVF International Conference on
  Computer Vision (ICCV)}, pp.\  4195--4205, October 2023.

\bibitem[Raffel et~al.(2020)Raffel, Shazeer, Roberts, Lee, Narang, Matena,
  Zhou, Li, and Liu]{JMLR:v21:20-074}
Colin Raffel, Noam Shazeer, Adam Roberts, Katherine Lee, Sharan Narang, Michael
  Matena, Yanqi Zhou, Wei Li, and Peter~J. Liu.
\newblock Exploring the limits of transfer learning with a unified text-to-text
  transformer.
\newblock \emph{Journal of Machine Learning Research}, 21\penalty0
  (140):\penalty0 1--67, 2020.

\bibitem[Shah et~al.(2026)Shah, Li, Bello, Zhu, and
  Martín-Martín]{ShahR-RSS-26}
Rutav Shah, Yisu Li, Femi Bello, Yuke Zhu, and Roberto Martín-Martín.
\newblock {Memory Retrieval in Visuomotor Policies for Long-Horizon Robot
  Control}.
\newblock In \emph{Proceedings of Robotics: Science and Systems}, Sydney,
  Australia, July 2026.
\newblock \doi{10.15607/RSS.2026.XXII.010}.

\bibitem[Sridhar et~al.(2026)Sridhar, Pan, Sharma, and Finn]{ICLR2026_9da515b1}
Ajay Sridhar, Jennifer Pan, Satvik Sharma, and Chelsea Finn.
\newblock Scaling up memory for robotic control via experience retrieval.
\newblock In \emph{International Conference on Learning Representations},
  volume 2026, pp.\  97142--97166, 2026.

\bibitem[{Team Wan} et~al.(2025){Team Wan}, Wang, Ai, Wen, Mao, Xie, Chen, Yu,
  Zhao, Yang, Zeng, Wang, Zhang, Zhou, Wang, Chen, Zhu, Zhao, Yan, Huang, Feng,
  Zhang, Li, Wu, Chu, Feng, Zhang, Sun, Fang, Wang, Gui, Weng, Shen, Lin, Wang,
  Wang, Zhou, Wang, Shen, Yu, Shi, Huang, Xu, Kou, Lv, Li, Liu, Wang, Zhang,
  Huang, Li, Wu, Liu, Pan, Zheng, Hong, Shi, Feng, Jiang, Han, Wu, and
  Liu]{wan2025}
{Team Wan}, Ang Wang, Baole Ai, Bin Wen, Chaojie Mao, Chen-Wei Xie, Di~Chen,
  Feiwu Yu, Haiming Zhao, Jianxiao Yang, Jianyuan Zeng, Jiayu Wang, Jingfeng
  Zhang, Jingren Zhou, Jinkai Wang, Jixuan Chen, Kai Zhu, Kang Zhao, Keyu Yan,
  Lianghua Huang, Mengyang Feng, Ningyi Zhang, Pandeng Li, Pingyu Wu, Ruihang
  Chu, Ruili Feng, Shiwei Zhang, Siyang Sun, Tao Fang, Tianxing Wang, Tianyi
  Gui, Tingyu Weng, Tong Shen, Wei Lin, Wei Wang, Wei Wang, Wenmeng Zhou, Wente
  Wang, Wenting Shen, Wenyuan Yu, Xianzhong Shi, Xiaoming Huang, Xin Xu, Yan
  Kou, Yangyu Lv, Yifei Li, Yijing Liu, Yiming Wang, Yingya Zhang, Yitong
  Huang, Yong Li, You Wu, Yu~Liu, Yulin Pan, Yun Zheng, Yuntao Hong, Yupeng
  Shi, Yutong Feng, Zeyinzi Jiang, Zhen Han, Zhi-Fan Wu, and Ziyu Liu.
\newblock Wan: Open and advanced large-scale video generative models, 2025.

\bibitem[Wu et~al.(2024)Wu, Shentu, Yi, Lin, and Abbeel]{wu2024gello}
Philipp Wu, Yide Shentu, Zhongke Yi, Xingyu Lin, and Pieter Abbeel.
\newblock {GELLO:} {A} general, low-cost, and intuitive teleoperation framework
  for robot manipulators.
\newblock In \emph{{IEEE/RSJ} International Conference on Intelligent Robots
  and Systems, {IROS} 2024, Abu Dhabi, United Arab Emirates, October 14-18,
  2024}, pp.\  12156--12163. {IEEE}, 2024.
\newblock \doi{10.1109/IROS58592.2024.10801581}.

\bibitem[Yang et~al.(2026)Yang, Mu, Wei, Lu, Li, Xu, Xue, Yuan, Lin, Pang, and
  Xu]{yang2026memorywam}
Sizhe Yang, Juncheng Mu, Tianming Wei, Chenhao Lu, Xiaofan Li, Linning Xu,
  Zhengrong Xue, Zhecheng Yuan, Dahua Lin, Jiangmiao Pang, and Huazhe Xu.
\newblock Memorywam: Efficient world action modeling with persistent memory,
  2026.

\bibitem[Ye et~al.(2026)Ye, Ge, Zheng, Gao, Yu, Kurian, Indupuru, Tan, Zhu,
  Xiang, Malik, Lee, Liang, Ranawaka, Gu, Xu, Wang, Hu, Narayan, Bjorck, Wang,
  Kim, Niu, Zheng, Xie, Wu, Wang, Julian, Xu, Du, Chebotar, Reed, Kautz, Zhu,
  Fan, and Jang]{ye2026dreamzero}
Seonghyeon Ye, Yunhao Ge, Kaiyuan Zheng, Shenyuan Gao, Sihyun Yu, George
  Kurian, Suneel Indupuru, You~Liang Tan, Chuning Zhu, Jiannan Xiang, Ayaan
  Malik, Kyungmin Lee, William Liang, Nadun Ranawaka, Jiasheng Gu, Yinzhen Xu,
  Guanzhi Wang, Fengyuan Hu, Avnish Narayan, Johan Bjorck, Jing Wang, Gwanghyun
  Kim, Dantong Niu, Ruijie Zheng, Yuqi Xie, Jimmy Wu, Qi~Wang, Ryan Julian,
  Danfei Xu, Yilun Du, Yevgen Chebotar, Scott Reed, Jan Kautz, Yuke Zhu,
  Linxi~"Jim" Fan, and Joel Jang.
\newblock World action models are zero-shot policies, 2026.
\newblock URL \url{https://arxiv.org/abs/2602.15922}.

\bibitem[Yuan et~al.(2026)Yuan, Dong, Liu, and Zhao]{yuan2026fastwam}
Tianyuan Yuan, Zibin Dong, Yicheng Liu, and Hang Zhao.
\newblock Fast-wam: Do world action models need test-time future imagination?,
  2026.

\bibitem[Zhao et~al.(2026)Zhao, Zhang, Su, Wang, Li, Yang, Tay, Jr., and
  Zhu]{zhao2026sgwamselfguidedworldmodeling}
Ruiteng Zhao, Zhengshen Zhang, Yue Su, Wenshuo Wang, Jiahui Li, Zhiyuan Yang,
  Francis E.~H. Tay, Marcelo H.~Ang Jr., and Haiyue Zhu.
\newblock {SG-WAM}: Self-guided world modeling in geometry-aware policy space,
  2026.

\bibitem[Zhu et~al.(2025)Zhu, Yu, Feng, Burchfiel, Shah, and Gupta]{zhu2025uwm}
Chuning Zhu, Raymond Yu, Siyuan Feng, Benjamin Burchfiel, Paarth Shah, and
  Abhishek Gupta.
\newblock {Unified World Models: Coupling Video and Action Diffusion for
  Pretraining on Large Robotic Datasets}.
\newblock In \emph{Proceedings of Robotics: Science and Systems}, LosAngeles,
  CA, USA, June 2025.
\newblock \doi{10.15607/RSS.2025.XXI.015}.

\end{thebibliography}
\bibliographystyle{iclr2027_conference}

\appendix
\clearpage
\section*{Appendix}
This appendix provides additional details on the implementation,
experimental settings, quantitative analyses, qualitative results, and
discussion to complement the main paper.

\begin{itemize}
    \item 
    \textbf{Section \ref{app:implementation_details}} describes the implementation details of TaskWAM, including model architectures, task-state token construction,
    training procedures, and inference settings.

    \item
    \textbf{Section \ref{app:experimental_details}} provides detailed experimental settings,
    including StageBench construction, RoboTwin 2.0 configurations,
    and real-world evaluation protocols.

    \item
    \textbf{Section \ref{app:quantitative}} presents supplementary quantitative analyses,
    including detailed benchmark results, additional ablations, and
    further analysis of task-state grounding.

    \item
    \textbf{Section \ref{app:qualitative}} provides supplementary qualitative results,
    including additional multi-stage task executions, stage confusion
    cases, and real-world manipulation demonstrations.

    \item
    \textbf{Section \ref{app:discussion}} discusses additional insights into the motivation behind TaskWAM.
\end{itemize}

\section{Implementation Details}
\label{app:implementation_details}
\subsection{Model Details}
\label{app:model_details}

\noindent\textbf{Notation.}
We use $t$ for a closed-loop decision step and $\tau$ for flow time.
The trajectory $\xi_i$ and stage sequence $\{s_i^1,\ldots,s_i^{n_i}\}$
follow Sec.~\ref{sec:problem_setup}.
The structured task state is
$\mathbf{T}_t=[\mathbf{M};\mathbf{S}_t]$, where the semicolon denotes
concatenation along the token dimension.
$N_m$ is the number of Mode Tokens, $K_{\mathrm{prior}}$ is the number
of codebook entries, and $K=N_s$ is the number of history slots.
We distinguish the state encoder's internal dimension $d_s=1024$
from the backbone conditioning dimension $d$.
Batch dimensions are omitted unless explicitly shown.

\subsubsection{Mode Tokens.}
\label{app:mode_tokens}

Reliable stage progression requires interpreting execution evidence
under the fine-grained requirements specified by the instruction.
When tasks share local interactions, even an accurate history may
support different continuations.
For example, after a microwave has started, one instruction may require
another preparation step during heating, whereas another requires
waiting until heating finishes.
The observed history can be identical, but the appropriate next stage
differs because of the timing requirement.
Execution evidence therefore needs to be considered together with the
instruction constraints that determine how the task should proceed.

We construct a persistent, instruction-grounded representation of
these constraints, which we refer to as \emph{Mode Tokens}.
Here, an execution mode describes how manipulation should proceed
under fine-grained requirements, including ordering, timing,
repetition, and arm assignment.
Mode Tokens represent these requirements, while State Tokens provide
evidence of the progress actually made.
Their complementary roles allow generation to use both the intended
execution requirements and the observed execution context.

The construction comprises two steps.
Instruction-dependent routing retrieves query slots from a learnable
codebook, providing a coarse execution-requirement prior.
Full-text grounding then refines these queries using the specific
requirements expressed in the current instruction.

\noindent\textbf{Instruction-Dependent Routing.}
Given a task instruction $\ell$, the text encoder produces
$\mathbf{C}=\mathcal{E}_{\mathrm{text}}(\ell)
\in\mathbb{R}^{L\times D}$,
where $L$ is the language-sequence length and $D$ is the text-feature
dimension.
We obtain a global instruction feature through masked average pooling:
\begin{equation}
\mathbf{c}_{\mathrm{pool}}
=
\frac{
\sum_{i=1}^{L}
\mathrm{mask}_{i}\mathbf{c}_{i}
}{
\sum_{i=1}^{L}\mathrm{mask}_{i}
},
\label{eq:app_mode_pool}
\end{equation}
where $\mathrm{mask}_{i}$ indicates a valid language token.
A two-layer MLP predicts routing logits and normalized weights:
\begin{equation}
\begin{aligned}
\boldsymbol{\ell}_{\mathrm{prior}}
&=
W_2\,
\mathrm{SiLU}
\left(
W_1\mathbf{c}_{\mathrm{pool}}+b_1
\right)
+b_2,\\
\boldsymbol{\alpha}
&=
\mathrm{softmax}
\left(
\boldsymbol{\ell}_{\mathrm{prior}}
\right)
\in\mathbb{R}^{K_{\mathrm{prior}}},
\end{aligned}
\label{eq:app_mode_routing}
\end{equation}
where $K_{\mathrm{prior}}$ is the number of codebook entries.
The weights determine an instruction-dependent combination of the
query slots used for subsequent requirement extraction.

\noindent\textbf{Learnable Requirement Codebook.}
We maintain a codebook
\begin{equation}
\mathbf{E}
\in
\mathbb{R}^{K_{\mathrm{prior}}\times N_m\times D},
\label{eq:app_mode_codebook}
\end{equation}
with $N_m$ learnable query slots per entry.
The routing weights combine corresponding slots across entries:
\begin{equation}
\mathbf{m}_{r}^{\mathrm{raw}}
=
\sum_{k=1}^{K_{\mathrm{prior}}}
\alpha_k\mathbf{E}_{k,r},
\qquad
r=1,\ldots,N_m.
\label{eq:app_mode_mix}
\end{equation}
The codebook parameters are shared across instructions, while their
combination adapts to the current language context.
The retrieved slots serve as a coarse execution-requirement prior
by initializing the queries that read the instruction.

This construction does not require a predefined one-to-one
correspondence between codebook entries and tasks.
Individual entries are not manually assigned to specific execution
requirements; their representations are learned jointly through the
world and action prediction objectives.
The retrieved mixture is subsequently refined against the full
instruction, so the final Mode Tokens depend on more than the routing
weights alone.

\noindent\textbf{Fine-Grained Instruction Grounding.}
Routing supplies an instruction-dependent query initialization,
but its pooled input does not preserve the full token-level structure
of the language sequence.
We therefore use the retrieved slots as queries over the complete
text context through a multi-layer Q-Former:
\begin{equation}
\begin{aligned}
\mathbf{M}^{\mathrm{raw}}
&=
[
\mathbf{m}_{1}^{\mathrm{raw}},
\ldots,
\mathbf{m}_{N_m}^{\mathrm{raw}}
],\\
\widetilde{\mathbf{M}}
&=
\mathcal{Q}_{\mathrm{ground}}
\left(
\mathbf{M}^{\mathrm{raw}},
\mathbf{C}
\right).
\end{aligned}
\label{eq:app_mode_grounding}
\end{equation}
The language tokens provide the key-value context, giving the queries
access to the words and contextual relationships that specify
ordering, timing, repetition, and arm assignment.
Grounding thus adapts the retrieved prior to the detailed requirements
of the current instruction.

We project the grounded queries to the backbone conditioning dimension:
\begin{equation}
\mathbf{M}
=
\mathcal{P}_{m}
\left(
\widetilde{\mathbf{M}}
\right)
\in
\mathbb{R}^{N_m\times d},
\label{eq:app_mode_projection}
\end{equation}
where $\mathcal{P}_{m}$ denotes the output projection.
In the implementation described here, each codebook entry contains
$N_m=8$ query slots, and the module outputs eight Mode Tokens with
dimension $d=4096$.
The number of query slots $N_m$ is distinct from the number of
codebook entries $K_{\mathrm{prior}}$.

\noindent\textbf{Joint Learning for World-Action Generation.}
We jointly train the routing head, codebook, grounding module, and
WAM using the backbone's world and action prediction objectives,
without a separate codebook pretraining stage.
As described in Sec.~\ref{sec:tffs}, Mode Tokens participate in
task-state attention, layer-wise modulation, and residual prediction.
These pathways allow the instruction-grounded representations to
influence both world and action generation.

Prediction losses backpropagate through the conditioning pathways
to the Mode-Token construction modules.
The learned queries are therefore optimized according to their
contribution to downstream world-action prediction, connecting
instruction grounding with the generative execution objective.

\noindent\textbf{Persistent Conditioning During Execution.}
Mode Tokens are retained while the instruction remains unchanged,
keeping the task-specific execution requirements accessible across
stage transitions.
State Tokens are updated from actual action--observation interactions,
providing evolving evidence of execution progress.
Both generation streams access these complementary representations
to condition subsequent world-action predictions.

At inference time, the routing weights are predicted from the task
instruction.
Constructing Mode Tokens does not require a user-provided task
identifier or codebook index.

\subsubsection{State Tokens.}
\label{app:state_encoder}

Mode Tokens specify how the task should proceed, but do not indicate
which requirements have already been fulfilled.
When similar observations recur across stages, execution progress must
be inferred from the preceding interactions.
An executed action records what the robot attempted, while the subsequent
observation and its change provide evidence of the resulting outcome.
We therefore construct State Tokens from temporally ordered
action--observation transitions, providing execution evidence for
subsequent world--action generation.

\noindent\textbf{Interaction Representation.}
We use a fixed history window of $K=8$ interaction slots.
Each slot $k$ pairs the preceding executed action chunk
$\mathbf{A}_{k-1}^{\mathrm{exec}}$ with the resulting visual latent $\mathbf{F}_k$
and its change
$\Delta\mathbf{F}_k=\mathbf{F}_k-\mathbf{F}_{k-1}$.
These inputs provide complementary evidence:
$\mathbf{A}_{k-1}^{\mathrm{exec}}$ describes the executed behavior,
$\mathbf{F}_k$ describes the resulting scene, and
$\Delta\mathbf{F}_k$ captures the observed change between interactions.
Their joint representation allows the encoder to use both the
executed actions and their visual feedback when tracking progress.

We independently project the three modalities into a 256-dimensional
feature space.
The same learnable slot-position embedding is added to all modality
features belonging to slot $k$, associating them with the same
interaction.

\noindent\textbf{Within-Slot Aggregation.}
A 2-layer, 8-head Q-Former aggregates the multimodal features within
each interaction slot.
Its query is formed as
\begin{equation}
\mathbf{q}_k
=
\mathbf{q}_{\mathrm{base}}+\mathbf{p}^{q}_k,
\end{equation}
where $\mathbf{q}_{\mathrm{base}}$ is a shared learnable query and
$\mathbf{p}^{q}_k$ is a learnable embedding for the slot position.
The Q-Former attends to the projected visual, visual-change, and
action features.
Its output is mapped to a 1024-dimensional interaction representation
$\mathbf{e}_k$, summarizing the executed behavior and the observed
scene information associated with that interaction.

\noindent\textbf{Causal Temporal Encoding.}
The eight interaction representations are arranged in execution order
and processed by a 2-layer, 8-head causal Transformer.
A lower-triangular attention mask restricts each slot to its own and
preceding interactions.
This preserves the chronology needed to distinguish similar scene
states reached through different execution histories.

The encoder outputs
\begin{equation}
\mathbf{S}^{\mathrm{enc}}_t
\in
\mathbb{R}^{B\times 8\times 1024},
\end{equation}
where $B$ denotes the batch size.
We retain all eight temporally ordered tokens, allowing subsequent
generation to access representations associated with different
positions in the recent interaction history.
Each later token can incorporate a longer prefix of the available
window.
A backbone-specific projection maps these representations to the
conditioning dimension $d$:
\begin{equation}
\mathbf{S}_t
=
\mathcal{P}_{s}
\left(\mathbf{S}^{\mathrm{enc}}_t\right)
\in
\mathbb{R}^{B\times 8\times d}.
\end{equation}

\noindent\textbf{History Initialization.}
At the beginning of an episode, fewer than eight interaction slots
may be available.
We pad missing slots with the earliest available observation and set
their action inputs to zero.
For the first valid interaction slot, the latent-difference input is
also set to zero because no preceding observation is available.
This keeps the input length fixed without carrying history across
episode boundaries.

\noindent\textbf{Closed-Loop State Updates.}
After executing action chunk $\mathbf{A}_t^{\mathrm{exec}}$, the robot receives the
next observation and encodes its visual latent $\mathbf{F}_{t+1}$.
The new interaction
$(\mathbf{F}_{t+1},\Delta\mathbf{F}_{t+1},\mathbf{A}_t^{\mathrm{exec}})$
is appended to the history window, the oldest slot is removed when
necessary, and the State Tokens are recomputed.
Subsequent generation therefore uses execution evidence updated from
the robot's actual interactions, alongside the persistent requirements
represented by Mode Tokens.

\subsubsection{Task-State Conditioning and Generation Steering.}

The grounded task state
$\mathbf{T}_t=[\mathbf{M};\mathbf{S}_t]$
provides both task requirements and execution evidence for world-action
generation.
TaskWAM retains the backbone's native world--action information flow and injects
$\mathbf{T}_t$ into both world and action streams through two
complementary mechanisms.

First, we append the task state to the key-value context of both streams,
allowing future world and action tokens to directly access the current
execution requirements and progress.
In addition, pooled Mode Tokens provide layer-wise AdaLN modulation:
\begin{equation}
    [\boldsymbol{\gamma}_{l}^{m},\boldsymbol{\beta}_{l}^{m}]
    =g_l(\bar{\mathbf{m}}),
    \qquad
    \widetilde{\mathbf{H}}_l
    =
    (1+\boldsymbol{\gamma}_{l}^{m})\odot
    \operatorname{LN}(\mathbf{H}_l)
    +\boldsymbol{\beta}_{l}^{m}.
    \label{eq:app_mode_adaln}
\end{equation}
Here, $\bar{\mathbf{m}}$ is the pooled Mode-Token representation and
$g_l$ predicts the modulation parameters for layer $l$.

Second, we introduce task-state-conditioned residual correction to
directly adjust the predicted world-action evolution.
For flow-based backbones and stream $b\in\{w,a\}$, the residual is generated by querying
the task state with the corresponding hidden states:
\begin{equation}
    \Delta\mathbf{v}_{t,\tau}^{b}
    =
    \mathcal{R}_b(\mathbf{H}_{t,\tau}^{b};\mathbf{T}_t),
    \qquad
    \widehat{\mathbf{v}}_{t,\tau}^{b}
    =
    \mathbf{v}_{t,\tau}^{b}
    +
    \gamma_b\Delta\mathbf{v}_{t,\tau}^{b},
    \label{eq:app_future_steering}
\end{equation}
where $\gamma_b$ is a learnable bounded scale and
$\mathcal{R}_b$ denotes a stream-specific cross-attention residual head.
The residual and base prediction have matching shapes.
For Cosmos Policy, the residual operates on denoised latent predictions
at the native EDM noise level, rather than on a flow velocity.
The two prediction parameterizations use the corresponding backbone
training objective, as described in Sec.~\ref{app:training_inference}.

\subsection{Training and Inference Details}
\label{app:training_inference}

\begin{table}[t]
    \centering
    \caption{Training configuration on kitchen manipulation tasks of StageBench.}
    \label{tab:libero_training}
    \small
    \begin{tabular}{@{}ll@{}}
        \toprule
        Hyperparameter & Setting \\
        \midrule
        Backbone & Wan2.2-TI2V-5B + FastWAM Action DiT \\
        Training GPUs & $16 \times$ NVIDIA H100 80GB \\
        Effective batch size & 128 \\
        Optimizer updates & 58,720 \\
        Optimizer & AdamW \\
        Learning rate & $1 \times 10^{-4}$ \\
        Learning-rate schedule & Cosine decay \\
        Warm-up ratio & 5\% \\
        Weight decay & 0.01 \\
        Gradient clipping norm & 1.0 \\
        Precision & BF16 \\
        Random seed & 42 \\
        Image views & Two $224 \times 224$ views \\
        Concatenated image resolution & $224 \times 448$ \\
        Video training window & 33 frames \\
        Action-to-video frequency ratio & 4 \\
        Action / proprioception dimensions & 7 / 8 \\
        History length & 8 interaction slots \\
        History observation-anchor interval & 16 action steps \\
        Executed-action chunk length & 16 steps \\
        \bottomrule
    \end{tabular}
\end{table}

\noindent\textbf{Training.}
TaskWAM is implemented on Fast-WAM in PyTorch, retaining the backbone's native inference structure.
Following the open-source training setup of Fast-WAM\cite{yuan2026fastwam}, Table~\ref{tab:libero_training} summarizes the training configuration
for long-horizon kitchen manipulation tasks constructed in LIBERO for StageBench.
The simulation experiments are trained on 16 NVIDIA H100 80GB GPUs with an
effective batch size of 128 for 10 epochs.
Flow-based WAMs use their flow-matching generative objectives,
whereas Cosmos Policy uses EDM denoising. 
The world and action
generative terms are denoted
$\mathcal{L}_{\mathrm{world}}+\mathcal{L}_{\mathrm{action}}$.
The routing head, codebook, grounding module, and WAM are trained jointly in one stage.

\noindent\textbf{Closed-loop inference.}
For a fixed instruction, Mode Tokens are computed once and retained
throughout the rollout. The policy uses the instruction, current visual
observation, and causal visual--action history to condition world and
action generation. After an action chunk is executed and the next
observation is encoded, the new visual--action transition is appended
to the history and the State Tokens are refreshed for the
next decision step. Only the executed actions, denoted by
$\mathbf{A}_t^{\mathrm{exec}}$, are stored as interaction evidence;
unexecuted predictions are not treated as completed actions.
Task-state conditioning and residual prediction follow each backbone's
native world--action information flow.

\section{Experimental Details}
\label{app:experimental_details}

\subsection{StageBench}
\label{app:stage_bench}

\noindent\textbf{Task families and instruction-defined variants.}
The eight StageBench task families detailed here comprise four
single-arm LIBERO families and four bimanual RoboTwin 2.0 families.
Task conditions specify subtask order, arm assignment, execution timing,
or repetition count. We refer to these instruction-defined conditions
as \emph{task variants}; they are distinct from the learned codebook entries.
The described variants contain up to five execution stages.
Table~\ref{tab:stage_bench_tasks} summarizes the task families and
their execution requirements.

\begin{table}[!htbp]
\centering
\caption{Task families and execution requirements in StageBench.}
\label{tab:stage_bench_tasks}
\small
\setlength{\tabcolsep}{4pt}
\renewcommand{\arraystretch}{1.10}
\begin{tabularx}{\linewidth}{@{}l p{0.38\linewidth} X@{}}
\toprule
ID & Task family & Execution requirements \\
\midrule
\multicolumn{3}{@{}l}{\textbf{Kitchen Manipulation}} \\
K1 & Cooking \& Cheese
   & Interleaved; Sequential \\
K2 & Cooking \& Juice
   & Interleaved; Sequential \\
K3 & Cooking \& Cleanup
   & Interleaved; Sequential \\
K4 & Cooking \& Breakfast
   & Interleaved; Sequential \\
\midrule
\multicolumn{3}{@{}l}{\textbf{Bimanual Manipulation}} \\
B1 & Size-Order Cubes
   & Small-to-Large; Large-to-Small \\
B2 & Color-Order Cups
   & RGB; BGR; RBG; GRB \\
B3 & Arm-Assigned Hammering
   & Left Arm; Right Arm \\
B4 & Repeated Bread Cutting
   & Instruction-specified cutting count (three cuts in the illustrated condition) \\
\bottomrule
\end{tabularx}
\end{table}
\FloatBarrier

\begin{figure}[!htbp]
    \centering
    \includegraphics[width=\linewidth]{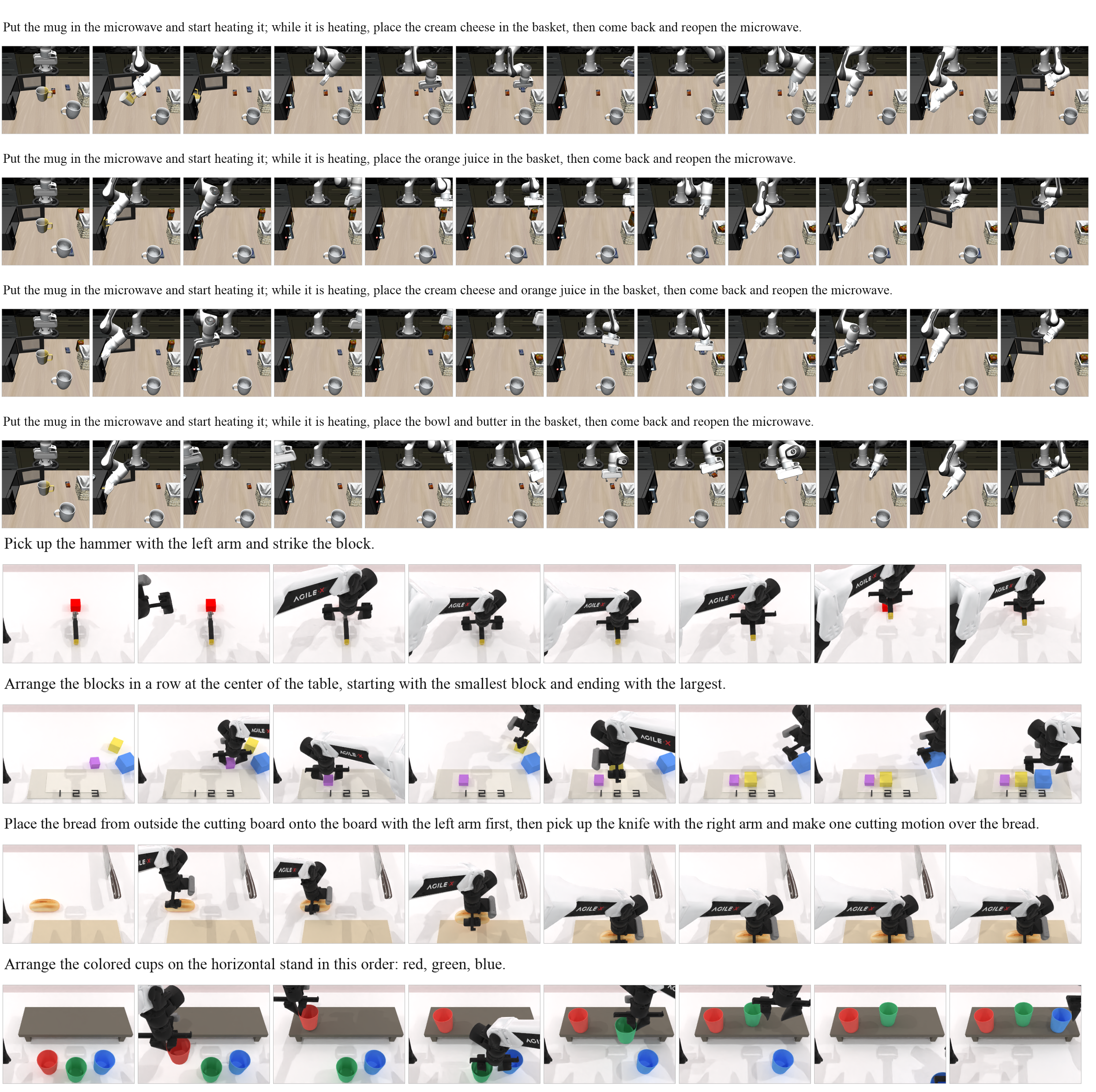}
    \caption{
    \textbf{StageBench benchmark visualization.}
    Representative task variants and their multi-stage execution processes
    are illustrated, highlighting diverse stage structures and instruction-
    dependent task requirements.
    }
    \label{fig:tasks}
\end{figure}

\noindent\textbf{StageBench Task Variants.}
The eight task families in Table~\ref{tab:stage_bench_tasks} instantiate
multiple task variants by modifying fine-grained instruction requirements.
Within a family, the conditions reuse the scene, objects, and local
manipulation primitives while specifying the constraints that determine
the correct continuation.
Figure~\ref{fig:tasks} visualizes representative prompts from each
family. The stage-aware evaluator distinguishes physical task completion
from instruction-defined execution routes based on task-specific events.

\noindent\textbf{Environment Extensions.}
To support realistic multi-stage manipulation and stage-aware evaluation,
we extend the LIBERO simulation environments with additional interactive
household assets.
For kitchen scenarios, we introduce functional microwave assets with
stateful behaviors, including door-triggered heating activation and
timer-based heating completion.
These extensions enable meaningful state transitions and temporal
dependencies beyond simple rearrangements of manipulation primitives.
All environment extensions and task trajectories are manually validated
over multiple rollouts to ensure reliable state transitions and
consistent event detection for automatic evaluation.

\begin{figure}[htbp]
    \centering
    \includegraphics[width=\linewidth]{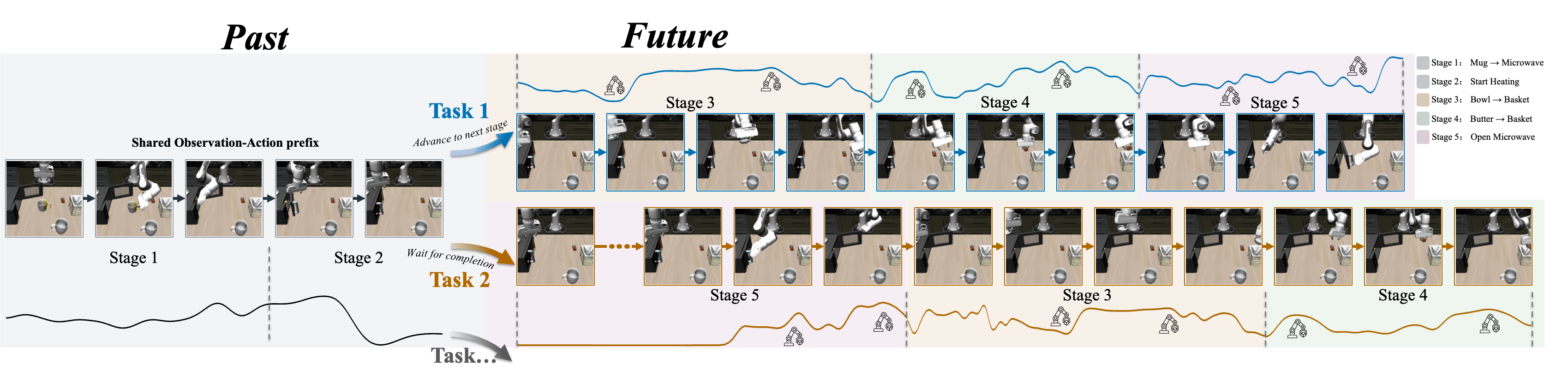}
    \caption{\textbf{Execution orders in a LIBERO household task.}
    Both variants share the mug-placement and heating prefix.
    The \emph{Interleaved} route collects the bowl and butter during heating and then reopens the microwave.
    The \emph{Sequential} route waits for heating to finish and reopens the microwave before collecting the objects.}
    \label{fig:libero}
\end{figure}

\medskip
\noindent\textbf{Microwave-based Cooking Tasks.}
These task families share the same interaction structure: placing the mug
into the microwave, starting heating, and performing a collection
operation. The variant variable is the temporal relation between
collection and microwave reopening, as shown in Figure~\ref{fig:libero}.
The \emph{Interleaved} prompt template is:
``Put the mug in the microwave and start heating it; while it is heating,
place \{target\} in the basket, then come back and reopen the microwave.''
The \emph{Sequential} template is:
``Put the mug in the microwave, wait for it to finish heating, reopen
the microwave, and only then place \{target\} in the basket.''
The target variable is instantiated as cream cheese, orange juice, cream
cheese and orange juice, or bowl and butter for the four cooking families.

\noindent\textbf{Size-Order Cubes.}
The task-variant variable is the required size order. The \emph{Small-to-Large}
prompt is ``Arrange the blocks in a row at the center of the table, starting
with the smallest block and ending with the largest.'' The \emph{Large-to-Small}
prompt is ``Arrange the blocks in a row at the center of the table, starting
with the largest block and ending with the smallest.'' Both variants arrange
the same blocks; evaluation checks their placement order against the instruction.

\begin{figure}[htbp]
    \centering
    \includegraphics[width=\linewidth]{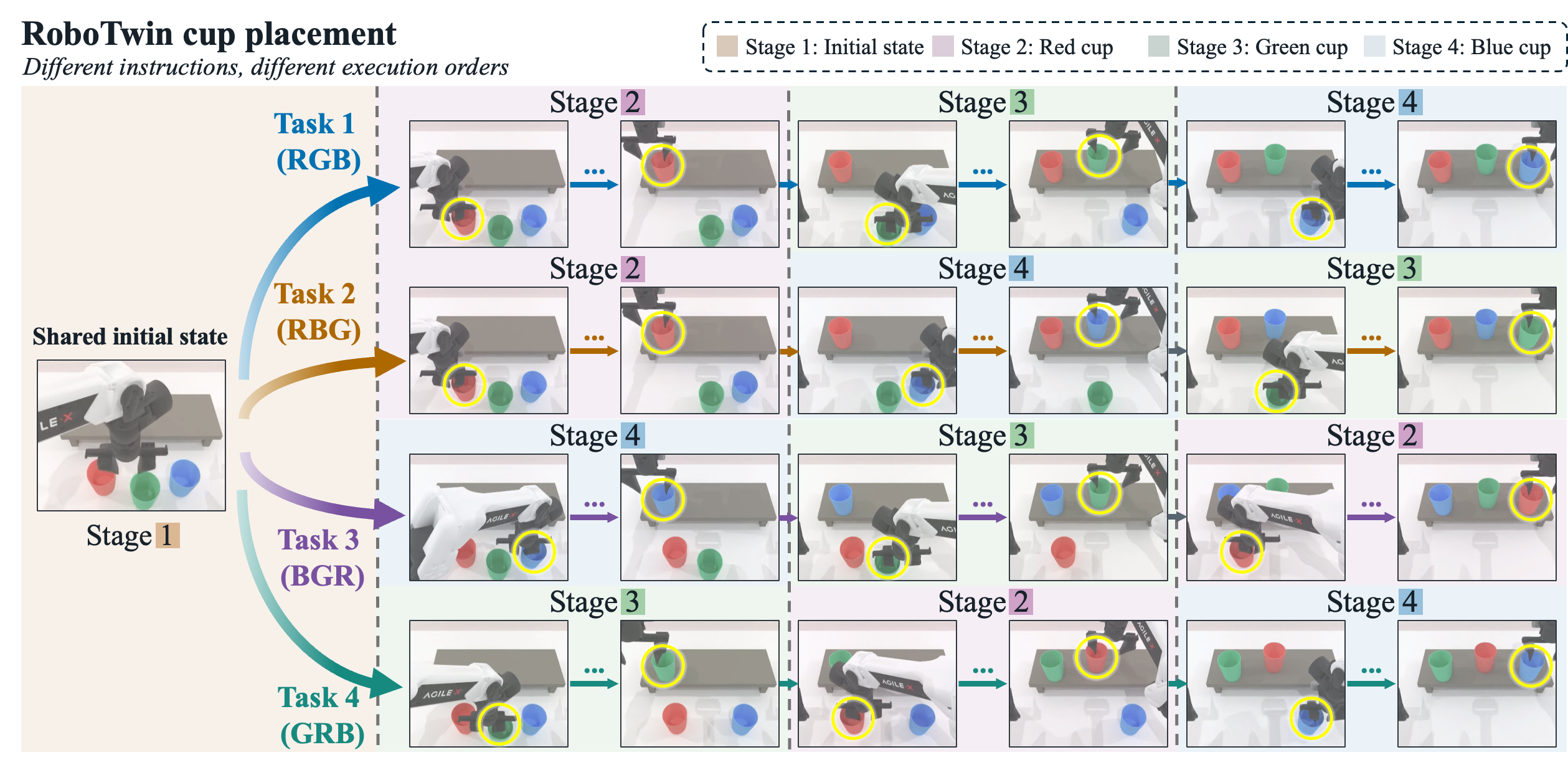}
    \caption{\textbf{Instruction-dependent cup placement in RoboTwin.}
    From a shared initial state, the robot places the red (R), green (G), and blue (B) cups in four prescribed orders: RGB, RBG, BGR, and GRB.
    Stage labels identify cup-specific operations, whose execution order varies with the instruction.}
    \label{fig:robotwin}
\end{figure}

\noindent\textbf{Color-Order Cups.}
The task-variant variable is the prescribed color sequence. For \emph{RGB},
the prompt is ``Arrange the colored cups on the horizontal stand in this order:
red, green, blue.'' For \emph{RBG}, it is ``Arrange the colored cups on the
horizontal stand in this order: red, blue, green.'' For \emph{BGR}, it is
``Arrange the colored cups on the horizontal stand in this order: blue, green,
red.'' For \emph{GRB}, it is ``Arrange the colored cups on the horizontal stand
in this order: green, red, blue.'' The variants share the initial scene and cup-placement
operations. Evaluation compares the observed color-specific placement sequence
with the requested prompt, as illustrated in Figure~\ref{fig:robotwin}.

\medskip
\noindent\textbf{Arm-Assigned Hammering.}
The variant variable is the instructed manipulation arm.
All variants share the same target object and manipulation procedure.
The prompt template is:
``Pick up the hammer with the \{arm\} arm and strike the block.''
where \{arm\} is instantiated as the left or right arm.

\medskip
\noindent\textbf{Repeated Bread Cutting.}
The task requirement is the specified repetition count.
The prompt is:
``Use both robot arms to make exactly three cuts in the bread.''
The evaluator checks whether the robot performs the required number of
cuts before termination.

\FloatBarrier

\noindent\textbf{Data collection.}
For the LIBERO tasks, we collect demonstrations through 3D human
teleoperation. For the four RoboTwin task families, we use the official
data generation pipeline. We collect 50 demonstration trajectories
for each collected instruction-defined condition. The resulting dataset contains 2.20 million frames
across both platforms, with approximately 12 hours spent on human
teleoperation.

\noindent\textbf{Event-based stage annotation.}
To support stage-level evaluation, we record task-relevant simulator
states at every environment step and convert them into timestamped
task events. The events are defined separately for each task family
according to its execution requirements. For example, the LIBERO
heating tasks record heating start, heating completion, post-heating
microwave reopening, target-object manipulation, and whether each
target object has entered the basket. A stage may depend on multiple
events rather than a single simulator condition.

For RoboTwin, the recorded events depend on the task family, such as
object placement order for B1 and B2, hammer manipulation by the
instructed arm for B3, and completed cutting motions for B4.
These event traces provide a common interface for tracking execution
progress while retaining task-specific stage definitions.

\noindent\textbf{Automatic evaluation.}
We separately evaluate physical task completion and instruction-level
execution consistency.
Physical success is determined using the native success criteria of the
corresponding simulation platform.
Meanwhile, deterministic finite-state machines analyze recorded event
sequences to reconstruct execution routes and track stage completion.
Stage-consistent task completion additionally requires satisfying the
instruction-defined ordering, timing, arm, and repetition constraints.

For the LIBERO heating tasks, let
$u_{\mathrm{collect}}$ denote the environment step at which the collection
branch is executed and $u_{\mathrm{open}}$ the environment step at which the
microwave is reopened after heating.
An execution is classified as \emph{Interleaved} when
$u_{\mathrm{collect}} < u_{\mathrm{open}}$, and as
\emph{Sequential} when
$u_{\mathrm{open}} < u_{\mathrm{collect}}$.
Heating-start and heating-completion events are additionally retained
to verify the corresponding physical stages and route consistency.

For RoboTwin, the evaluator applies task-specific rules to verify
instruction-defined constraints, including object ordering, arm roles,
and repetition counts.

\noindent\textbf{Stage Progression Rate.}
To complement final task success, we report Stage Progression Rate
(SPR), which measures whether a policy completes the instructed
next stage after completing the current stage.
For task family $f$ and variant $\nu$, let $\mathcal{E}_{f,\nu}$ denote
the set of eligible transition opportunities across evaluated
rollouts. An opportunity becomes eligible once the preceding
required stages have been completed in order.
Let $\mathcal{C}_{f,\nu}\subseteq\mathcal{E}_{f,\nu}$ contain opportunities
for which the first recognizable subsequent stage entry matches
the instructed next stage and that stage is completed before
rollout termination. We define
\begin{equation}
\mathrm{SPR}_{f,\nu}
=
\frac{|\mathcal{C}_{f,\nu}|}{|\mathcal{E}_{f,\nu}|}
\times 100\%.
\label{eq:stage_bench_spr}
\end{equation}
The denominator retains all eligible opportunities, including
incorrect stage entries, correct entries without completion,
and cases with no recognizable subsequent entry.
Task-level SPR pools these counts across variants before division.
Higher SPR indicates more reliable progression along the
instructed execution route.
Because unsuccessful opportunities also include ordinary manipulation
failures, SPR measures stage-consistent progression; event traces are
used to identify specific wrong-stage entries.

\noindent\textbf{Evaluation protocol.}
For each LIBERO task family, the two task variants are evaluated using
the same fixed set of 100 unique initial states.
Each task--variant pair is therefore evaluated on 100 initial states,
and all compared models use the same initial states and the same
task-specific maximum number of environment steps.
For every rollout, we save the detected events and their time steps,
physical task success, identified execution route, and stage-level
progress.

\FloatBarrier
\subsection{RoboTwin 2.0 Setup}
\label{app:robotwin50}

\noindent\textbf{Benchmark.}
We evaluate on all 50 bimanual manipulation tasks in RoboTwin 2.0 \citep{chen2025robotwin2} under the Clean and Randomized settings. 
The latter introduces variations in scene appearance and configuration to assess robustness to environmental changes. 
The Clean training set contains 2,500 demonstrations (50 per task), and the Randomized training set contains 25,000 demonstrations
(500 per task). Simulation experiments are conducted over 100 trials.

\noindent\textbf{Configuration.}
For this benchmark, TaskWAM augments Fast-WAM \citep{yuan2026fastwam}
with Structured Task-State Grounding and Task-State Conditioned Future Steering.
Mode Tokens provide instruction-grounded execution requirements, while
State Tokens summarize causal action--observation transitions.
Both token sets condition the world and action streams through attention
and residual prediction; pooled Mode Tokens also provide layer-wise
modulation. These additions follow the task-state pathways defined in
Sec.~\ref{sec:tffs} while retaining Fast-WAM's native world--action
information flow.

\subsection{Real-World Setup}
\label{app:real_world}

\noindent\textbf{Platforms and data collection.}
The real-world platforms include a single-arm Franka Research 3,
dual-arm AgileX Piper (6-DoF) and AgileX Nero (7-DoF), and a Unitree G1
humanoid equipped with BrainCo Revo2 dexterous hands.
Franka is equipped with third-person and wrist-mounted RealSense D435
cameras, while the dual-arm platforms use wrist cameras (and a
head-mounted camera for Nero) for visual observations.
The humanoid platform is operated via PICO VR for whole-body
teleoperation.
Teleoperation interfaces include Gello \cite{wu2024gello} for gripper-based robots,
TypeTele \cite{lin2025typetelereleasingdexterityteleoperation} for dexterous hands, and SONIC \cite{sonic} for the humanoid platform.
We collect 100--150 teleoperated demonstrations per task.

\noindent\textbf{Multi-Stage Real-World Tasks.}
We evaluate four multi-stage tasks spanning three representative real-world scenarios.
These tasks evaluate cross-task and within-task stage confusion under shared objects, repeated interactions, and instruction-dependent continuations.
\ding{172} \textbf{Sequential Cup Nesting} requires nesting red, pink, and blue cups in the instructed order, with three order variants.
The breakfast-preparation scenario includes two tasks: \ding{173} \textbf{Bread Toasting \& Spreading}, which requires toasting bread, transferring it to a plate, and applying the specified blueberry or yellow-peach jam; and \ding{174} \textbf{Bread Cutting}, which requires making exactly three cuts and then stopping.
\ding{175} \textbf{Laboratory Fluid Manipulation} requires transferring a reagent into the receiving container specified by the instruction.
These settings naturally reuse objects and local interactions across tasks and execution stages, providing real-world evaluation of both cross-task and within-task stage confusion.
Additionally, we evaluate Bread Toasting \& Spreading under a modified
tabletop background to examine robustness against scene variations.

\noindent\textbf{(1) Sequential Cup Nesting.}
The task-variant variable is the prescribed color sequence.
All variants share the same scene and manipulation procedure. The prompt
template is:
``Assemble the colored cups by nesting them one by one on the tabletop
platform in the following color order: \{color order\}, and place the
assembled cup stack into the basket.''
where \{color order\} is instantiated as \emph{pink, red, blue} (PRB),
\emph{red, blue, pink} (RBP), and \emph{red, pink, blue} (RPB).

\begin{figure}[!htbp]
    \centering
    \includegraphics[width=\linewidth]{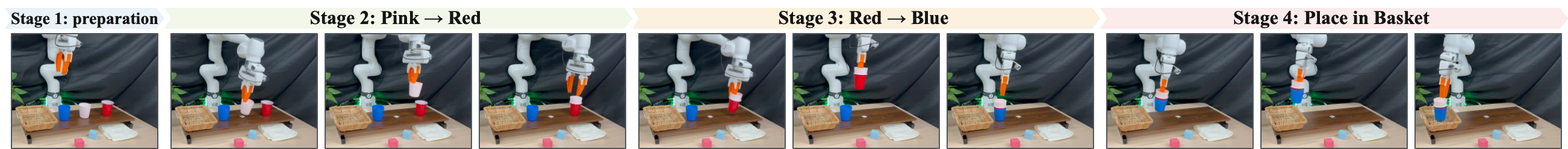}
    \caption{\textbf{Visualization of Sequential Cup Nesting.}}
    \label{fig:real_cup_nesting}
\end{figure}
\FloatBarrier

\noindent\textbf{(2) Bread Toasting \& Spreading.}
The task-variant variable is the specified spreading ingredient.
All variants share the same preparation procedure. The prompt template is:
``Heat the bread in the toaster, take it out when toasted, place it on a
plate, and spread \{ingredient\} jam.''
where \{ingredient\} is instantiated as \emph{blueberry} and
\emph{yellow peach}.

\begin{figure}[!htbp]
    \centering
    \includegraphics[width=\linewidth]{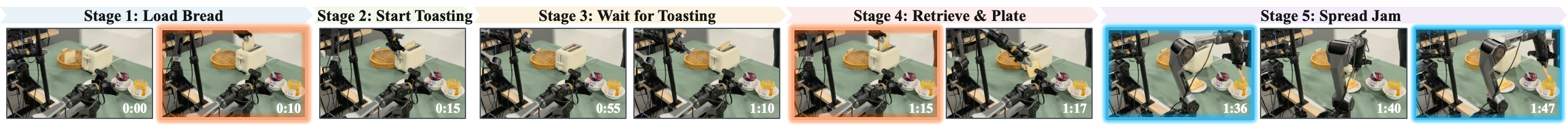}
    \caption{\textbf{Visualization of Bread Toasting \& Spreading.}}
    \label{fig:real_bread_toasting}
\end{figure}
\FloatBarrier

\Needspace{11\baselineskip}
\noindent\textbf{(3) Laboratory Fluid Manipulation.}
The task-variant variable is the receiving container.
Both variants share the same manipulation sequence. The prompt template is:
``Use the right arm to remove the chemical liquid container from the
microwave and pour its contents into the \{container\} for collection,
then use the left arm to close the microwave door.''
where \{container\} is instantiated as \emph{Petri dish} and
\emph{gel plate}.

\begin{figure}[!htbp]
    \centering
    \includegraphics[width=\linewidth]{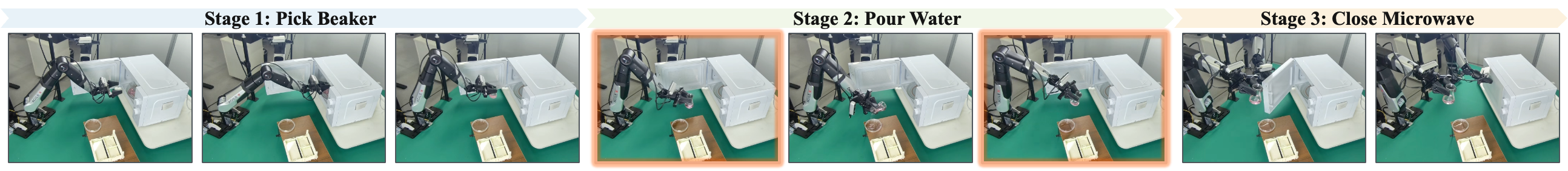}
    \caption{\textbf{Visualization of Laboratory Fluid Manipulation.}}
    \label{fig:real_lab}
\end{figure}
\FloatBarrier

\noindent\textbf{(4) Bread Cutting.}
The task requirement is the repetition count.
The prompt is:
``Use both robot arms to make exactly three cuts in the bread.''

\begin{figure}[!htbp]
    \centering
    \includegraphics[width=\linewidth]{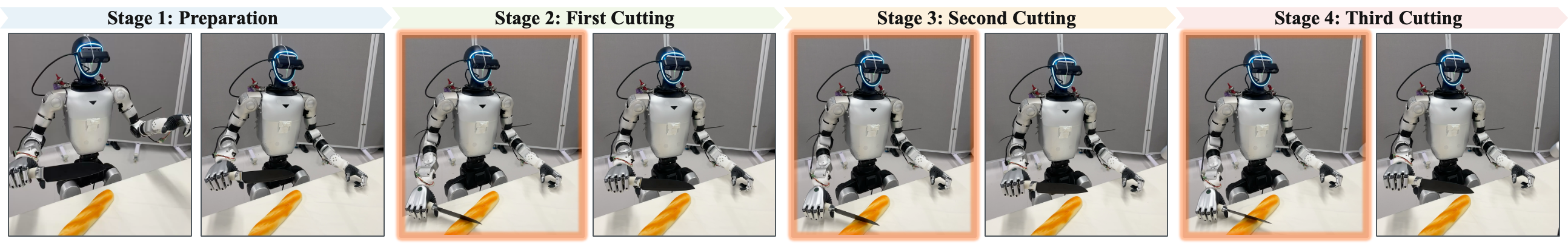}
    \caption{\textbf{Visualization of Bread Cutting.}}
    \label{fig:real_bread_cutting}
\end{figure}
\FloatBarrier

\noindent\textbf{Evaluation.}
Cumulative stage completion and overall task success are defined
as successful trials divided by total attempts. A trial counts
as completing stage $k$ only after it has completed the required
sequence through that stage. The evaluation pools count across the variants within each task family.
Each task contains 1-3 variants, with each variant evaluated over 10-20 trials.

\section{Supplementary Quantitative Analysis}
\label{app:quantitative}

\subsection{Detailed Results}
\label{app:detailed_results}

The complete RoboTwin 2.0 task-wise comparison under the Clean and Randomized settings is reported in Table~\ref{tab:robotwin50_taskwam_fastwam_full}.

\begin{table}[!h]
\caption{\textbf{Complete RoboTwin 2.0 task-wise comparison.}}
\label{tab:robotwin50_taskwam_fastwam_full}
\centering
\setlength{\tabcolsep}{3.2pt}
\renewcommand{\arraystretch}{0.84}
\resizebox{\textwidth}{!}{%
\begin{tabular}{@{}l cc|cc|cc|cc|cc|cc|cc@{}}
\toprule
\multirow{2}{*}{\textbf{Task}} & \multicolumn{2}{c|}{Motus} & \multicolumn{2}{c|}{$\pi_{0.5}$} & \multicolumn{2}{c|}{LingBot-VA} & \multicolumn{2}{c|}{LaWAM} & \multicolumn{2}{c|}{AHA-WAM} & \multicolumn{2}{c|}{Fast-WAM} & \multicolumn{2}{c@{}}{\textbf{w/ TaskWAM (ours)}} \\
\cmidrule(lr){2-3}\cmidrule(lr){4-5}\cmidrule(lr){6-7}\cmidrule(lr){8-9}\cmidrule(lr){10-11}\cmidrule(lr){12-13}\cmidrule(l){14-15}
& Clean & Rand. & Clean & Rand. & Clean & Rand. & Clean & Rand. & Clean & Rand. & Clean & Rand. & Clean / \scriptsize $\Delta$ & Rand. / \scriptsize $\Delta$ \\
\midrule
Adjust Bottle & 89 & 93 & \textbf{100} & 99 & 90 & 94 & \textbf{100} & \textbf{100} & \textbf{100} & \textbf{100} & \textbf{100} & \textbf{100} & \textbf{100} & \textbf{100} \\
Beat Block Hammer & 95 & 88 & 96 & 93 & 96 & \textbf{98} & 90 & 93 & \textbf{100} & 94 & 99 & 97 & \textbf{100} & \textbf{98} \\
Blocks Ranking RGB & 99 & 97 & 92 & 85 & 99 & 98 & 97 & \textbf{100} & \textbf{100} & 98 & \textbf{100} & \textbf{100} & \textbf{100} & 99 \\
Blocks Ranking Size & 75 & 63 & 49 & 26 & 94 & 96 & 93 & 89 & \textbf{98} & \textbf{98} & 94 & \textbf{98} & 93 & 93 \\
Click Alarmclock & \textbf{100} & \textbf{100} & 98 & 89 & 99 & \textbf{100} & \textbf{100} & \textbf{100} & \textbf{100} & \textbf{100} & \textbf{100} & \textbf{100} & \textbf{100} & \textbf{100} \\
Click Bell & \textbf{100} & \textbf{100} & 99 & 66 & \textbf{100} & \textbf{100} & \textbf{100} & \textbf{100} & \textbf{100} & \textbf{100} & \textbf{100} & \textbf{100} & \textbf{100} & \textbf{100} \\
Dump Bin Bigbin & 95 & 91 & 92 & \textbf{97} & 89 & 96 & 97 & 95 & \textbf{99} & 92 & 97 & 96 & 97 & 96 \\
Grab Roller & \textbf{100} & \textbf{100} & \textbf{100} & \textbf{100} & \textbf{100} & \textbf{100} & \textbf{100} & \textbf{100} & \textbf{100} & \textbf{100} & \textbf{100} & \textbf{100} & \textbf{100} & \textbf{100} \\
Handover Block & 86 & 73 & 66 & 57 & \textbf{99} & 78 & 96 & \textbf{87} & 96 & 85 & 95 & 81 & 82 & 86 \\
Handover Mic & 78 & 63 & 98 & 97 & 94 & 96 & 93 & 98 & 99 & 99 & 99 & \textbf{100} & \textbf{100} & 99 \\
Hanging Mug & 38 & 38 & 18 & 17 & 40 & 28 & 51 & 43 & \textbf{85} & \textbf{78} & 58 & 62 & 55 & 61 \\
Lift Pot & 96 & 99 & 96 & 85 & \textbf{100} & 99 & \textbf{100} & 99 & \textbf{100} & \textbf{100} & \textbf{100} & \textbf{100} & \textbf{100} & \textbf{100} \\
Move Can Pot & 34 & 74 & 51 & 55 & 94 & \textbf{97} & \textbf{98} & 93 & 78 & 83 & 90 & 88 & 95 & 96 \\
Move Pillbottle Pad & 93 & 96 & 84 & 61 & 99 & \textbf{99} & 97 & 90 & \textbf{100} & \textbf{99} & \textbf{100} & \textbf{99} & 98 & \textbf{99} \\
Move Playingcard Away & \textbf{100} & 96 & 96 & 84 & \textbf{100} & 99 & \textbf{100} & \textbf{100} & \textbf{100} & \textbf{100} & \textbf{100} & \textbf{100} & \textbf{100} & \textbf{100} \\
Move Stapler Pad & 83 & 85 & 56 & 42 & 91 & 79 & \textbf{94} & \textbf{87} & 88 & 77 & 77 & 64 & 82 & 75 \\
Open Laptop & 95 & 91 & 90 & 96 & 92 & 94 & \textbf{100} & \textbf{100} & 98 & 97 & 98 & \textbf{100} & \textbf{100} & \textbf{100} \\
Open Microwave & \textbf{95} & \textbf{91} & 34 & 77 & 82 & 86 & 41 & 43 & 40 & 46 & 62 & 45 & 79 & 77 \\
Pick Diverse Bottles & 90 & \textbf{91} & 81 & 71 & 89 & 82 & \textbf{91} & 88 & 89 & 85 & 80 & 85 & 81 & 83 \\
Pick Dual Bottles & 96 & 90 & 93 & 63 & \textbf{100} & \textbf{99} & \textbf{100} & 95 & 98 & \textbf{99} & \textbf{100} & 96 & 86 & 90 \\
Place A2B Left & 88 & 79 & 87 & 82 & 97 & 93 & 98 & 91 & \textbf{100} & 92 & 95 & 93 & 97 & \textbf{97} \\
Place A2B Right & 91 & 87 & 87 & 84 & \textbf{97} & 95 & 89 & 94 & 96 & 92 & 93 & \textbf{99} & 96 & 95 \\
Place Bread Basket & 91 & 94 & 77 & 64 & 97 & 95 & 92 & 85 & 89 & 94 & 91 & 93 & \textbf{98} & \textbf{97} \\
Place Bread Skillet & 86 & 83 & 85 & 66 & \textbf{95} & 90 & 90 & 83 & 91 & 87 & 90 & 93 & 94 & \textbf{98} \\
Place Burger Fries & 98 & 98 & 94 & 87 & 97 & 95 & 93 & 96 & \textbf{99} & 98 & 96 & \textbf{99} & \textbf{99} & 97 \\
Place Can Basket & 81 & 76 & 62 & 62 & 81 & 84 & \textbf{92} & 65 & 81 & \textbf{94} & 71 & 69 & 72 & 80 \\
Place Cans Plasticbox & 98 & 94 & 94 & 84 & \textbf{100} & 99 & \textbf{100} & 95 & \textbf{100} & \textbf{100} & 99 & 96 & 99 & 97 \\
Place Container Plate & 98 & 99 & 99 & 95 & 99 & 97 & \textbf{100} & \textbf{100} & \textbf{100} & 96 & 96 & \textbf{100} & 99 & 99 \\
Place Dual Shoes & 93 & 87 & 75 & 75 & 94 & 89 & \textbf{98} & 94 & 91 & \textbf{99} & 94 & 88 & 93 & 86 \\
Place Empty Cup & 99 & 98 & \textbf{100} & 99 & \textbf{100} & \textbf{100} & 99 & \textbf{100} & \textbf{100} & \textbf{100} & \textbf{100} & \textbf{100} & \textbf{100} & \textbf{100} \\
Place Fan & 91 & 87 & 87 & 85 & 99 & 93 & 92 & 93 & \textbf{100} & 87 & 96 & \textbf{96} & 98 & 94 \\
Place Mouse Pad & 66 & 68 & 60 & 39 & \textbf{93} & \textbf{96} & 91 & 84 & \textbf{93} & 82 & 83 & 89 & \textbf{93} & \textbf{96} \\
Place Object Basket & 81 & 87 & 80 & 76 & 91 & 88 & \textbf{92} & \textbf{90} & 81 & 80 & 89 & 88 & \textbf{92} & 89 \\
Place Object Scale & 88 & 85 & 86 & 80 & \textbf{96} & 95 & 95 & 88 & \textbf{96} & 87 & 90 & 97 & \textbf{96} & \textbf{99} \\
Place Object Stand & 98 & \textbf{97} & 91 & 85 & \textbf{99} & 96 & 92 & 93 & 94 & 93 & 90 & 94 & 96 & 93 \\
Place Phone Stand & 87 & 86 & 81 & 81 & 97 & 97 & 93 & 94 & \textbf{98} & 96 & 97 & \textbf{99} & 90 & 98 \\
Place Shoe & 99 & 97 & 92 & 93 & 98 & 98 & \textbf{100} & \textbf{100} & 91 & 96 & 96 & 99 & 88 & 98 \\
Press Stapler & 93 & 98 & 87 & 83 & 85 & 82 & \textbf{98} & 97 & \textbf{98} & \textbf{99} & 90 & 97 & 95 & 93 \\
Put Bottles Dustbin & 81 & 79 & 84 & 79 & 87 & 91 & 94 & 92 & 83 & \textbf{96} & \textbf{95} & 90 & 88 & 90 \\
Put Object Cabinet & 88 & 71 & 80 & 79 & 85 & 87 & 90 & 82 & 91 & 89 & \textbf{94} & 89 & 90 & \textbf{90} \\
Rotate QRcode & 89 & 73 & 89 & 87 & \textbf{96} & \textbf{91} & 94 & 89 & 90 & 90 & 93 & 89 & 87 & \textbf{91} \\
Scan Object & 67 & 66 & 72 & 65 & \textbf{96} & 91 & \textbf{96} & 90 & 94 & 90 & 89 & \textbf{92} & 95 & 89 \\
Shake Bottle & \textbf{100} & 97 & 99 & 97 & \textbf{100} & 97 & \textbf{100} & \textbf{100} & \textbf{100} & \textbf{100} & \textbf{100} & \textbf{100} & \textbf{100} & \textbf{100} \\
Shake Bottle Horizontally & \textbf{100} & 98 & 99 & 99 & \textbf{100} & 99 & \textbf{100} & \textbf{100} & \textbf{100} & \textbf{100} & \textbf{100} & \textbf{100} & \textbf{100} & \textbf{100} \\
Stack Blocks Three & 91 & 95 & 91 & 76 & \textbf{99} & \textbf{98} & 90 & 75 & \textbf{99} & \textbf{98} & 95 & 97 & 95 & \textbf{98} \\
Stack Blocks Two & \textbf{100} & 98 & 97 & \textbf{100} & \textbf{100} & 98 & \textbf{100} & 97 & \textbf{100} & 99 & \textbf{100} & \textbf{100} & \textbf{100} & \textbf{100} \\
Stack Bowls Three & 79 & \textbf{87} & 77 & 71 & 86 & 83 & \textbf{90} & 80 & 79 & 82 & 80 & 81 & 87 & 79 \\
Stack Bowls Two & 98 & 98 & 95 & 96 & 94 & 98 & \textbf{100} & 99 & \textbf{100} & \textbf{100} & 92 & 98 & 97 & 96 \\
Stamp Seal & 93 & 92 & 79 & 55 & 96 & \textbf{97} & 89 & 88 & \textbf{98} & 90 & 90 & 94 & 80 & 88 \\
Turn Switch & \textbf{84} & \textbf{78} & 62 & 54 & 44 & 45 & 47 & 56 & 70 & 74 & 61 & 59 & 73 & 74 \\
\midrule
\rowcolor{taskwamgainrow}
\textbf{Average (\%)} & 88.66 & 87.02 & 82.74 & 76.76 & 92.90 & 91.50 & 92.52 & 89.48 & \textbf{93.40} & 92.20 & 91.88 & 91.78 & 92.70 \scriptsize\taskwamgain{(+0.82)} & \textbf{93.06} \scriptsize\taskwamgain{(+1.28)} \\
\bottomrule
\end{tabular}%
}
\end{table}

\subsection{Additional Ablations and Analysis}
\label{app:additional_analysis}

\par
\leavevmode\begin{wraptable}[8]{r}{0.35\columnwidth}
\centering
\caption{\textbf{Analysis of State Tokens.}}
\label{tab:state_structure}
\scriptsize
\resizebox{\linewidth}{!}{%
\begin{tabular}{l|c}
\toprule
\textbf{Setting}
& \textbf{\textit{Suc.}$\uparrow$}\\
\midrule
TaskWAM
& \textbf{69.4}\\
\hline
w/ Shuffled Temporal Order
& 50.3\\
w/ Repeated Latest State
& 58.1\\
\bottomrule
\end{tabular}%
}
\end{wraptable}
\noindent\textbf{Effectiveness of temporally structured State Tokens.}
Table~\ref{tab:state_structure} examines whether State Tokens
benefit from structured execution history beyond the latest state.
Shuffling token order preserves their content but destroys chronology, reducing success from 69.4\% to 50.3\%; repeating the latest token removes state evolution and yields 58.1\%.
These results support the contribution of temporal order and historical
state evolution to the full model's task-execution performance.
\par

\subsection{Training Convergence Analysis}
\label{app:training_convergence}

\begin{figure}[t]
    \centering
    \includegraphics[width=0.48\linewidth]{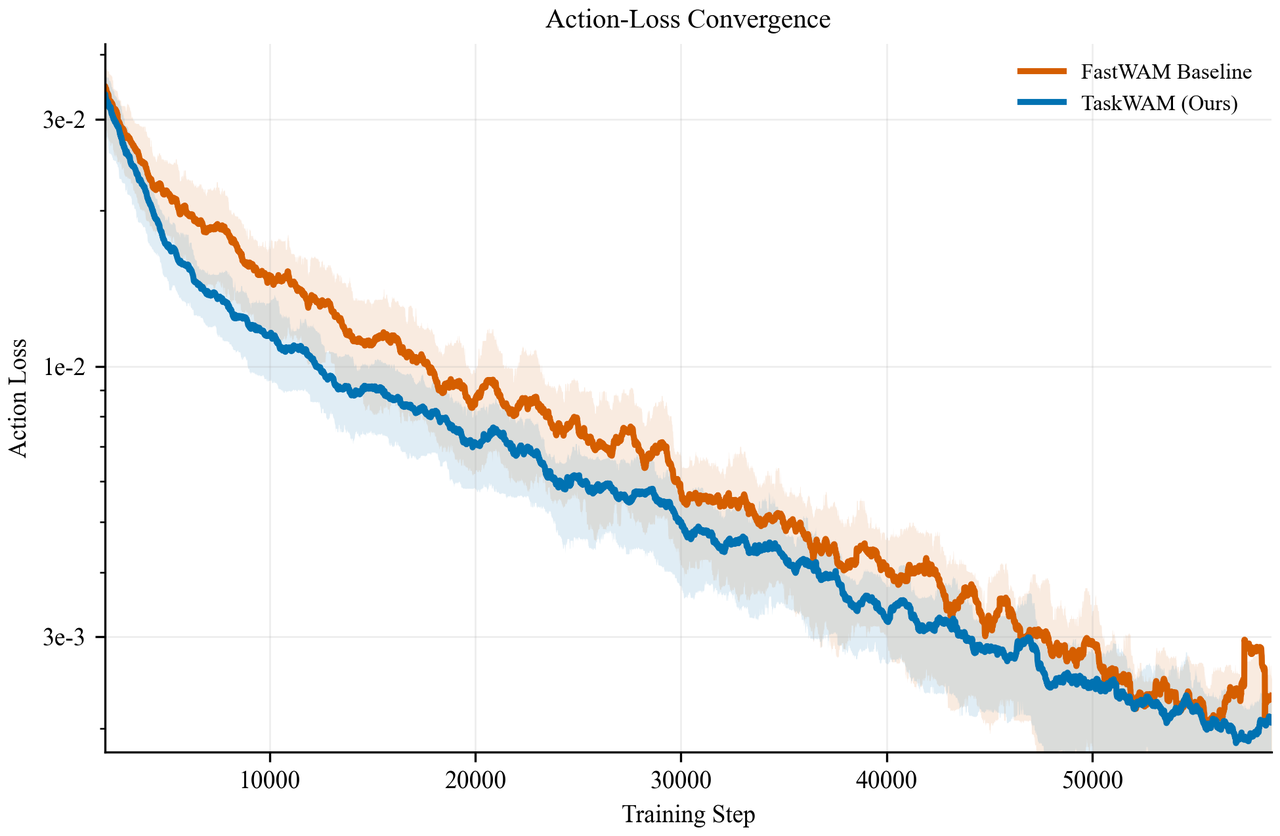}
    \caption{\textbf{Training convergence.}
    TaskWAM converges faster and reaches a lower action loss than FastWAM under the same training setup.}
    \label{fig:action_convergence}
\end{figure}

As shown in Figure~\ref{fig:action_convergence}, explicitly grounding world-action generation in structured task states enables faster convergence and a lower action loss than FastWAM.
The curves support improved action-loss convergence under the reported
training setup. They do not isolate which task-state component accounts
for the optimization difference; the component ablation in
Table~\ref{tab:component_ablation} evaluates their contributions to
execution performance.

\section{Supplementary Qualitative Analysis}
\label{app:qualitative}

\begin{figure}[!htbp]
    \centering
    \includegraphics[width=\linewidth]{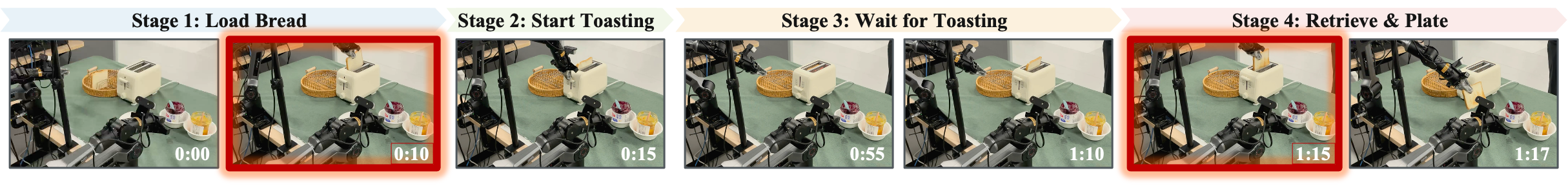}
    \caption{
    \textbf{Real-world stage confusion in Bread Toasting \& Spreading.}
    Stage 1 (loading bread) and Stage 4 (retrieving toasted bread) exhibit
    similar visual configurations, while Stage 3 introduces a long waiting
    period with redundant observations.
    Baseline WAMs may confuse the current progress with earlier stages,
    resulting in repeated or incorrect execution.
    }
    \label{fig:bread_case}
\end{figure}

\subsection{Real-World Stage Confusion Case Study}
\label{app:stage_confusion_case}

We further analyze a representative stage confusion case from the
real-world Bread Toasting \& Spreading task.
As shown in Figure~\ref{fig:bread_case}, the task contains a long waiting
stage during bread toasting, followed by a retrieval stage with a visual
configuration similar to the initial loading stage.

During the waiting period, the observation history contains many
visually repetitive frames.
At retrieval, a configuration resembling the loading stage must be
interpreted using the instruction and available interaction evidence.
In the illustrated baseline failure, the continuation returns to earlier
preparation behavior rather than completing retrieval, leading to
incorrect behaviors such as repeating the heating procedure or failing
to advance to subsequent stages.

TaskWAM conditions generation on instruction-grounded task
requirements from Mode Tokens with causal execution evidence from
State Tokens. The stage-wise results in Table~\ref{tab:real_world_results}
provide quantitative context for this qualitative comparison.

\subsection{Additional Qualitative Analysis}
Figures~\ref{fig:tasks}, \ref{fig:libero}, and \ref{fig:robotwin}
visualize the instruction-defined task variants and their corresponding
multi-stage execution structures in StageBench.
These examples illustrate how StageBench creates challenging yet
meaningful ambiguities by preserving shared scenes, objects, and local
interactions while varying task requirements and execution progress.

Figure~\ref{fig:tasks} presents representative task trajectories from
the benchmark task families, covering diverse single-arm and bimanual
manipulation scenarios.
Figure~\ref{fig:libero} shows LIBERO variants that share early-stage
operations, such as mug placement and microwave heating, but differ in
later-stage requirements, including whether the robot should continue
with collection or reopen the microwave after heating.
These variants create cross-task ambiguity where similar observations
require different continuations depending on the instruction.

Figure~\ref{fig:robotwin} presents four cup-placement variants from the
same initial configuration, where only the requested color order changes.
Although the manipulation primitives remain identical, the robot must
maintain the instruction-defined ordering constraint throughout the
execution.
These examples correspond to the event-order and stage-completion
checks described in Sec.~\ref{app:stage_bench}, and demonstrate how
StageBench evaluates reliable progression under overlapping
observation-action trajectories.

\section{Discussion}
\label{app:discussion}

\noindent\textbf{Implicit Stage Modeling in Vanilla WAMs.}
For a flow-based WAM, the world and action velocity fields are
conditioned on the instruction $\ell$, current observation $o_t$, and
optional history $\mathcal{H}_t$, and trained to match the target
velocities of demonstrated futures:
\begin{equation}
    \mathcal{L}_{\mathrm{wam}}
    =
    \mathbb{E}
    \Bigl[
        \sum_{b\in\{w,a\}}
        \bigl\|
        \mathbf{v}_{\theta}^{b}
        (\mathbf{x}_{t,\tau}^{b},\tau
        \mid \ell,o_t,\mathcal{H}_t)
        -
        \mathbf{u}_{t,\tau}^{b}
        \bigr\|_2^2
    \Bigr],
    \label{eq:base_flow_matching}
\end{equation}
where $b\in\{w,a\}$ denotes the world or action stream, $\tau$ is the
flow timestep, $\mathbf{x}_{t,\tau}^{b}$ is the corresponding flow input,
and $\mathbf{u}_{t,\tau}^{b}$ is the target velocity.

Demonstration losses already supervise the intended world and action
outputs. Local predictive plausibility alone, however, does not establish
compliance with the active task's stage requirements.
Under overlapping trajectories, observations and short histories can be
similar while the instruction or achieved progress calls for a different
continuation.
TaskWAM organizes these conditioning signals into instruction-grounded
Mode Tokens and interaction-derived State Tokens, then makes them
accessible through backbone conditioning and residual prediction.
This changes how task information enters generation while retaining
the backbone's generative training objective.

\noindent\textbf{Temporal Scope of Task-State Grounding.}
Mode Tokens retain instruction-grounded requirements across the rollout,
while State Tokens summarize the most recent $K$ interactions.
The resulting progress representation depends on evidence available
within this finite window.
The task state guides continuous world--action generation through learned
conditioning and residual prediction; the benchmark evaluator separately
uses symbolic event traces to assess stage compliance.

\end{document}